\documentclass[10pt,journal,cspaper,compsoc]{IEEEtran}
\ifCLASSOPTIONcompsoc
  \usepackage[nocompress]{cite}
\else
\fi
\ifCLASSINFOpdf
    \usepackage[pdftex]{graphicx}
    \DeclareGraphicsExtensions{.pdf,.jpeg,.png}
\else
    \usepackage[dvips]{graphicx}
    \DeclareGraphicsExtensions{.eps}
\fi
\usepackage[cmex10]{amsmath}
\usepackage{algorithmic}
\usepackage{url}

\usepackage{multirow}
\usepackage{algorithm}

\usepackage{color}
\usepackage{amsmath}
\usepackage{amssymb}
\usepackage{mathrsfs}
\usepackage{booktabs}
\usepackage{comment}
\usepackage{multirow}
\usepackage{makecell,rotating}
\usepackage{array}
\usepackage{setspace}
\usepackage{framed}
\usepackage{bbm}
\usepackage[dvipsnames, svgnames, x11names]{xcolor}

\usepackage{makecell}
\usepackage{multirow}

\begin{document}
%
\title{Quantifying the Knowledge in a DNN to Explain Knowledge Distillation for Classification}

%
%

\author{Quanshi Zhang\IEEEauthorrefmark{1}\IEEEauthorrefmark{2}, Xu Cheng\IEEEauthorrefmark{1}, Yilan Chen, and Zhefan Rao
\thanks{ \IEEEauthorrefmark{1} Contribute equally to this paper.}
\thanks{ \IEEEauthorrefmark{2} Correspondence.}
\IEEEcompsocitemizethanks{\IEEEcompsocthanksitem Quanshi Zhang (zqs1022@sjtu.edu.cu) is with the Department of Computer Science and Engineering,
the John Hopcroft Center, and the MoE Key Lab of Artificial Intelligence, AI Institute, at the Shanghai Jiao Tong University, China. }
}

%
%

\markboth{IEEE TRANSACTIONS ON PATTERN ANALYSIS AND MACHINE INTELLIGENCE}%
{Shell \MakeLowercase{\textit{et al.}}: Bare Demo of IEEEtran.cls for Computer Society Journals}
%


\IEEEcompsoctitleabstractindextext{%

\begin{abstract}
Compared to traditional learning from scratch, knowledge distillation sometimes makes the DNN achieve superior performance.
In this paper, we provide a new perspective to explain the success of knowledge distillation based on the information theory, \textit{i.e.} quantifying knowledge points encoded in intermediate layers of a DNN for classification.
To this end, we consider the signal processing in a DNN as a layer-wise process of discarding information.
A knowledge point is referred to as an input unit, the information of which is discarded much less than that of other input units.
Thus, we propose three hypotheses for knowledge distillation based on the quantification of knowledge points.
\textbf{1.} The DNN learning from knowledge distillation encodes more knowledge points than the DNN learning from scratch. 
\textbf{2.} Knowledge distillation makes the DNN more likely to learn different knowledge points simultaneously. 
In comparison, the DNN learning from scratch tends to encode various knowledge points sequentially.
\textbf{3.} The DNN learning from knowledge distillation is often more stably optimized than the DNN learning from scratch.
To verify the above hypotheses, we design three types of metrics with annotations of foreground objects to analyze feature representations of the DNN, \textit{i.e.} the quantity and the quality of knowledge points, the learning speed of different knowledge points, and the stability of optimization directions. 
In experiments, we diagnosed various DNNs on different classification tasks, including image classification, 3D point cloud classification, binary sentiment classification, and question answering, which verified the above hypotheses.
\end{abstract}


}

\maketitle
\IEEEdisplaynotcompsoctitleabstractindextext
\IEEEpeerreviewmaketitle

\section{Introduction}
Knowledge distillation~\cite{hinton2015distilling} has achieved great success in various applications \cite{romero2014fitnets,yim2017gift,furlanello2018born}.
Knowledge distillation refers to the process of transferring the knowledge from one or more well-trained deep neural networks (DNNs), namely the teacher networks,
to a compact-yet-efficient DNN, namely the student network.
However, there have been very few studies to explain how and why knowledge distillation outperforms traditional learning from scratch.

Hinton~\textit{et al.}~\cite{hinton2015distilling} attributed the superior performance of knowledge distillation to the use of ``soft targets''.
Furlanello~\textit{et al.}~\cite{furlanello2018born} conjectured the effect of knowledge distillation on re-weighting training examples.
Lopez-Paz~\textit{et al.}~\cite{lopez2015unifying} interpreted knowledge distillation as a form of learning with privileged information. 
Besides, from the perspective of network training, knowledge distillation was regarded as a label-smoothing regularization method~\cite{tang2020understanding,yuan2020revisiting}, and was shown to enable faster convergence~\cite{yim2017gift,phuong2019towards}. 
These explanations of knowledge distillation were mainly based on qualitative statements or focused on the optimization and regularization effects.

Beyond the above perspectives for explanations, in this study, we provide a new perspective to analyze the success of knowledge distillation, \textit{i.e.} quantifying knowledge points encoded in the intermediate layer of a DNN. 
In other words, the quantity and the quality of knowledge points enable us to explain the representation power of the DNN, which is trained for classification.
For a better understanding, let us take the image classification as an example.
If a DNN learns lots of visual concepts for classification, then these concepts can be regarded as knowledge points, such as the bird head concept for the bird classification.
If these knowledge points are sufficiently discriminative, then this DNN probably achieves good performance.

\begin{figure*}[t]
    \centering
    \includegraphics[width=0.98\linewidth]{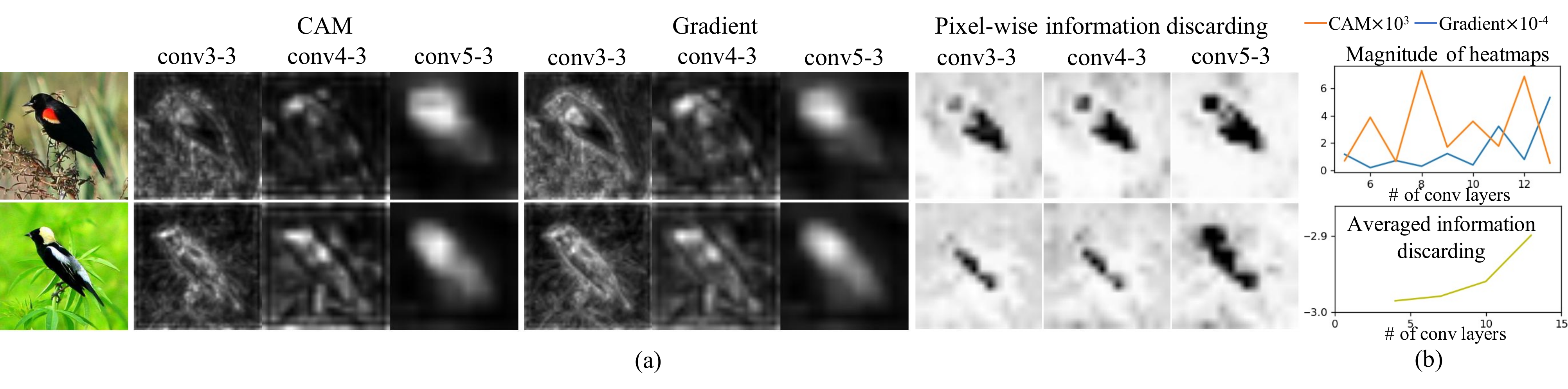}
    \vspace{-10pt}
    \caption{(a) Visualization of the information-discarding process through different layers of VGG-16 on the CUB200-2011 dataset. Input units with darker colors discard less information.
    (b) Comparison of the coherency of different explanation methods. 
    Our method coherently measures how input information is gradually discarded through layers, and enables fair comparisons of knowledge representations between different layers. In comparison, CAM~\cite{zhou2016learning} and gradient explanations~\cite{simonyan2017deep} cannot generate coherent magnitudes of importance values through different layers for fair comparisons. More analysis is presented in Section~\ref{hypothesis1}.}
    \vspace{-10pt}
    \label{fig:layer}
\end{figure*}
\begin{figure}[t]
    \centering
    \includegraphics[width=0.98\linewidth]{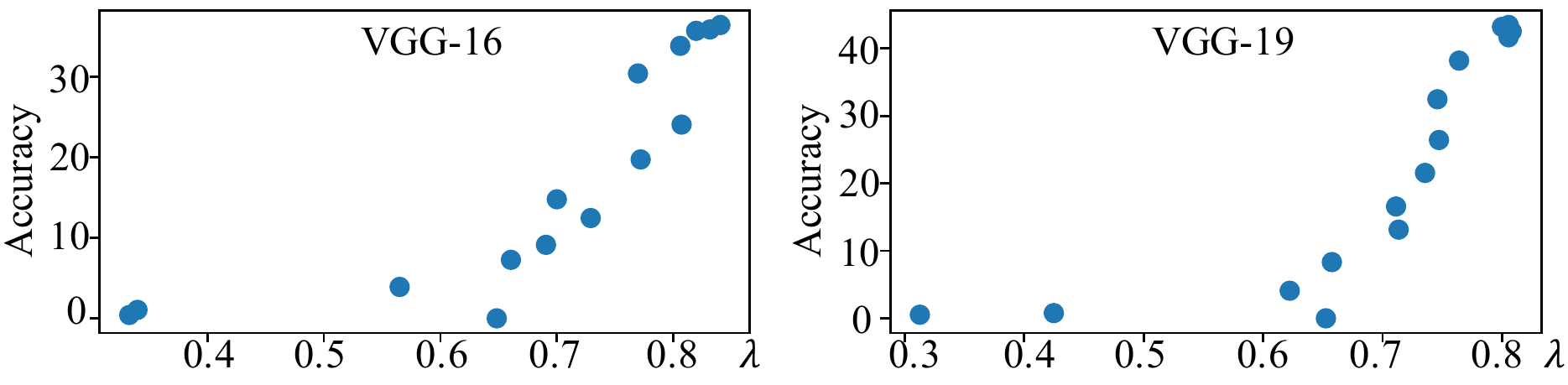}
    \vspace{-10pt}
    \caption{Positive correlation between the ratio of foreground knowledge points $\lambda$ and the classification accuracy of DNNs on the CUB200-2011 dataset.}
    \vspace{-15pt}
    \label{fig:acc_know}
\end{figure}

To measure the quantity and the quality of knowledge points encoded in a DNN, we consider the signal processing in the forward propagation of a DNN as a layer-wise process of discarding information.
As Figure~\ref{fig:layer}(a) shows, the DNN gradually discards the information of each input unit in the forward propagation. 
Here, an input unit is referred to as a variable (or a set of variables) in the input sample.
For example, the embedding of each word in the input sentence can be regarded as an input unit for natural language processing.
In the scenario of Figure~\ref{fig:layer}(a), an input unit is referred to as a pixel (or pixels within a small local region) of the input image.
Then, the information of some units (\textit{e.g.} information of pixels in the background) is significantly discarded.
In comparison, information of other input units (\textit{e.g.} information of pixels in the foreground) is discarded less.

In this way, we consider input units without significant information discarding as knowledge points, such as input units of the bird in Figure~\ref{fig:layer}(a). 
These knowledge points usually encode discriminative information for inferences,~\textit{e.g.} the wings of birds in Figure~\ref{fig:layer}(a) are discriminative for the fine-grained bird classification.
In comparison, input units with significant information-discarding usually do not contain sufficient discriminative information for prediction.
Thus, measuring the amount of information of input units discarded by the DNN enables us to quantify knowledge points.

Aside from the quantity of knowledge points, we also measure the quality of knowledge points with the help of annotations of foreground objects.
To this end, we classify all knowledge points into two types, including those related and unrelated to the task.
In this way, we can compute the ratio of knowledge points encoded in a DNN that are relevant to the task as a metric to evaluate the quality of knowledge points. 
A high ratio of relevant knowledge points indicates that the feature representation of a DNN is reliable.
Figure~\ref{fig:acc_know} illustrates that DNNs encoding a larger ratio of task-related knowledge points usually achieve better performance
(please see Section~\ref{4.2} for further analysis).

Based on the quantification of knowledge points, we propose the following three hypotheses as to why knowledge distillation outperforms traditional learning from scratch. 
Furthermore, we design three types of metrics to verify these hypotheses.

\textbf{Hypothesis 1: Knowledge distillation makes the DNN learn more knowledge points than the DNN learning from scratch.} 
The information-bottleneck theory~\cite{wolchover2017new,shwartz2017opening} claimed that a DNN tended to preserve features relevant to the task and neglected those features irrelevant to the task for inference.
We assume that a teacher network is supposed to achieve superior classification performance, because the teacher network usually has a more complex network architecture or is often learned using more training data.
In this scenario, the teacher network is considered to encode more knowledge points related to the task and fewer knowledge points unrelated to the task than the student network.
In this way, knowledge distillation forces the student network to mimic the logic of the teacher network, which ensures that the student network encodes more knowledge points relevant to the task and fewer knowledge points irrelevant to the task than the DNN learning from scratch.

\textbf{Hypothesis 2: Knowledge distillation makes the DNN more likely to learn different knowledge points simultaneously. In comparison, the DNN learning from scratch tends to encode various knowledge points sequentially through different epochs.} 
This is the case because knowledge distillation forces the student network to mimic all knowledge points of the teacher network simultaneously.
In comparison, the DNN learning from scratch tends to focus on different knowledge points in different epochs sequentially.

\textbf{Hypothesis 3: Knowledge distillation usually yields more stable optimization directions than learning from scratch.}
The DNN learning from scratch usually keeps trying different knowledge points in early epochs and then discarding unreliable ones in later epochs, rather than modeling target knowledge points directly.
Thus, the optimization directions of the DNN learning from scratch are inconsistent. 
For convenience, in this study, the phenomenon of inconsistent optimization directions through different epochs is termed as~\textbf{\textit{``detours''}}\footnote[1]{``Detours'' refers to the phenomenon that a DNN tends to encode various knowledge points in early epochs and discard non-discriminative ones later.}.

In contrast to learning from scratch, knowledge distillation usually forces the student network to mimic the teacher network directly.
The student network straightforwardly encodes target knowledge points without temporarily modeling and discarding other knowledge points.
Thus, the student network is optimized without significant detours.
For a better understanding of Hypothesis 3, let us take the fine-grained bird classification task as an example. With the guidance of the teacher network, the student network directly learns discriminative features from heads or tails of birds without significant detours. In comparison, the DNN learning from scratch usually extracts features from the head, belly, tail, and tree-branch in early epochs, and later neglects features from the tree branch. 
In this case, the optimization direction of the DNN learning from scratch is more unstable.

Note that previous studies on explanations of the optimization stability and the generalization power of the DNN were mainly conducted from the perspective of the parameter space.
\cite{simsekli2019tail,dinh2017sharp,keskar2016large} demonstrated that the flatness of minima of the loss function resulted in good generalization. 
\cite{neyshabur2015norm,NIPS2017_b22b257a,neyshabur2018a} illustrated that the generalization behavior of a network depended on the norm of the weights.
In contrast, this study explains the efficiency of optimization from the perspective of knowledge representations, \textit{i.e.} analyzing whether or not the DNN encodes different knowledge points simultaneously and whether they do so without significant detours.

\textbf{Connection to the information-bottleneck theory.}
We roughly consider that the information-bottleneck theory~\cite{wolchover2017new,shwartz2017opening} can also be used to explain the change of knowledge points in a DNN. It showed that a DNN usually extracted a few discriminative features for inferences, and neglected massive features irrelevant to inferences during the training process.
The information-bottleneck theory measures the entire discrimination power of each sample.
In comparison, our method quantifies the discrimination power of each input unit in a more detailed manner, and measures pixel-wise information discarded through forward propagation.

\textbf{Methods.} 
To verify the above hypotheses, we design three types of metrics to quantify knowledge points encoded in intermediate layers of a DNN, and analyze how different knowledge points are modeled during the training procedure.

$\bullet\quad$The first type of metrics measures the quantity and quality of knowledge points.
We consider that a well-trained DNN with superior classification performance is supposed to encode a massive amount of knowledge points, most of which are discriminative for the classification of target categories.

$\bullet\quad$The second type of metrics measures whether a DNN encodes various knowledge points simultaneously or sequentially, \textit{i.e.} whether different knowledge points are learned at similar speeds.

$\bullet\quad$The third metric evaluates whether a DNN is optimized with or without significant detours.
In other words, this metric measures whether or not a DNN directly learns target knowledge points without temporarily modeling and discarding other knowledge points.
To this end, this metric is designed as an overlap between knowledge points encoded by the DNN in early epochs and the final knowledge points modeled by the DNN.
If this overlap is large, then we consider this DNN is optimized without significant detours; otherwise, with detours.
This is the case because a small overlap indicates that 
the DNN first encodes a large number of knowledge points in early epochs, but many of these knowledge points are discarded in later epochs.

In summary, we use these metrics to compare knowledge distillation and learning from scratch, in order to verify Hypotheses 1-3. 

\textbf{Whether to use the annotated knowledge points.}
In particular, we should define and quantify knowledge points without any human annotations.
It is because we do not want the analysis of the DNN to be influenced by manual annotations with significant subjective bias, which negatively impacts the trustworthiness of this research.
Besides, annotating all knowledge points is also typically expensive to the point of being impractical or impossible.
However, previous studies often define visual concepts (knowledge points) by human annotations, and these concepts usually have specific semantic meanings.
For example, Bau~\textit{et al}~\cite{bau2017network} manually annotated six types of semantic visual concepts (objects, parts, textures, scenes, materials, and colors) to explain feature representations of DNNs.
In this way, these methods cannot fairly quantify the actual amount of knowledge encoded by a DNN, because most visual concepts are not countable and cannot be defined with human annotations.

In contrast, in this study, we use the information in each input unit discarded by the DNN to define and quantify knowledge points without human annotations.
Such knowledge points are usually referred to as \textit{\textbf{``Dark Matters'' }}~\cite{xie2017learning}.
Compared to traditionally labeled visual concepts, we consider that these dark-matter knowledge points can provide a more general and trustworthy way to explain the feature representations of DNNs.

\textbf{Scope of explanation.}
A trustworthy explanation of knowledge distillation usually involves two requirements.
First, the teacher network should be well optimized.
Otherwise, if the teacher network does not converge or is optimized for other tasks, then the teacher network is not qualified to perform distillation.
Second, knowledge distillation often has two fully different utilities. 
(1) Distilling from high-dimensional features usually forces the student network to mimic all sophisticated feature representations in the teacher network.
(2) Distilling from the low-dimensional network output mainly selectively uses confident training samples for learning, 
with very little information on how the teacher encodes detailed features. 
In this case, the distillation selectively emphasizes the training of simple samples and ignores the training of hard samples\footnote[2]{Confident (simple) samples usually generate more salient output signals for knowledge distillation than unconfident (hard) samples.}.
Therefore, in this study, we mainly explain the first utility of knowledge distillation, 
\textit{i.e.,} forcing the student network to mimic the knowledge of the teacher network.
This utility is  mainly exhibited by distilling from high-dimensional intermediate-layer features.

Contributions of this study are summarized as follows.

1. The quantification of knowledge points encoded in intermediate layers of DNNs can be used as a new perspective to analyze DNNs.

2.  Based on the quantification of knowledge points, we design three types of metrics to explain the mechanism of knowledge distillation.

3. Three hypotheses about knowledge distillation are proposed and verified in different classification applications, including image classification, 3D point cloud classification, binary sentiment classification, and question answering,
which shed new light on the interpretation of knowledge distillation.

A preliminary version of this work appeared in \cite{cheng2020explaining}.

\section{Related work}
Although DNNs have shown considerable promises in various applications, they are still regarded as black boxes. 
Previous studies on interpreting DNNs can be roughly summarized as semantic explanations and mathematical analyses of the representation capacity.

$\bullet\quad$\textbf{Semantic explanations for DNNs.} 
An intuitive way to explain DNNs is to visualize the appearance encoded in intermediate layers of DNNs, which may significantly activate a specific neuron of a certain layer. 
Gradient-based methods~\cite{zhou2016learning,simonyan2017deep,yosinski2015understanding,mahendran2015understanding} measured the importance of intermediate-layer activation units or input units by using gradients of outputs \textit{w.r.t.} inputs. 
Inversion-based methods~\cite{dosovitskiy2016inverting}  inverted feature maps of convolutional  layers into images. 
Moreover, some studies estimated the pixel-wise or regional attribution/importance/saliency of an input to the network output~\cite{ribeiro2016should,selvaraju2017grad,kindermans2017learning,fong2017interpretable,chattopadhay2018grad,ribeiro2018anchors,Kapishnikov_2019_ICCV,hooker2019benchmark}.

The semantic dissection is also a classical direction of interpreting DNNs. Bau~\textit{et al.} ~\cite{bau2017network} manually annotated six types of visual concepts (\textit{objects, parts, scenes, textures, materials} and \textit{colors}), and associated units of DNN feature maps with these concepts. 
Furthermore, the first two concepts could be summarized as shapes, and the last four concepts as textures~\cite{zhanginterpretable}. 
Fong and Vedaldi~\cite{fong2018net2vec} used a set of filters to represent a single semantic concept, and built up many-to-many projections between concepts and filters. TCAV~\cite{kim2017interpretability} measured the importance of user-defined concepts to classification results.

Learning a DNN with interpretable feature representations in an unsupervised or weakly-supervised manner is another direction widely explored in explainable AI.
InfoGAN~\cite{chen2016infogan} and $\beta$-VAE~\cite{higgins2017beta} learned interpretable latent representations for generative networks. In the capsule network~\cite{sabour2017dynamic,hinton2018matrix}, each capsule encoded different meaningful features. 
Interpretable CNNs~\cite{zhanginterpretable} were used to make each filter in a high convolutional layer represent a specific object part.  
The BagNet~\cite{Brendel19iclr} was proposed to classify images based on the occurrences of local image features. 
The ProtoPNet~\cite{chen2019looks}  was introduced to learn prototypical parts.
Shen~\textit{et al.}~\cite{shen2021interpretable} proposed a method to modify a traditional CNN into an
interpretable compositional CNN.

In summary, previous studies have mainly been based on heuristic assumptions to calculate importance/saliency/attention~\cite{zeiler2014visualizing,yosinski2015understanding,mahendran2015understanding,simonyan2017deep} or used massive amounts of human-annotated concepts~\cite{bau2017network,kim2017interpretability} to interpret network features. 
In contrast, in this paper, we explain feature representations from the perspective of knowledge points.

$\bullet\quad$\textbf{Mathematical explanations of the representation capacity of DNNs.} 
Formulating and evaluating the representation capacity of DNNs is an emerging direction in research on explainable AI. 
The information-bottleneck theory~\cite{wolchover2017new,shwartz2017opening} provided a generic metric to quantify the information encoded in DNNs, which can be extended to evaluate the representation capacity of DNNs~\cite{goldfeld2019estimating,xu2017information} as well. 
Zhang~\textit{et al.} \cite{zhang2016understanding} discussed the relationship between the parameter number and the generalization capacity of DNNs. 
Weng~\textit{et al.}~\cite{weng2018evaluating} proposed to use the CLEVER score to measure the robustness of neural networks. 
Fort~\textit{et al.}~\cite{fort2019stiffness} used the stiffness to diagnose the generalization of a DNN.
The module criticality~\cite{Chatterji2020The} was proposed to explain the superior generalization performance of some network architectures.

In contrast to previous works, in this study, we aim to bridge the gap between mathematical explanations and semantic explanations. 
That is, we use the information discarding of the input to quantify knowledge points and to explain the discrimination power of the DNN. 
More specifically, we analyze the quantity and the quality of knowledge points in a DNN, measure whether a DNN encodes various knowledge points sequentially or simultaneously, and evaluate the stability of its optimization direction.

$\bullet\quad$\textbf{Knowledge distillation.} 
Knowledge distillation is a popular technique in knowledge transferring, which can be applied to different domains such as adversarial defense \cite{papernot2016distillation}, object detection~\cite{uijlings2018revisiting}, semantic segmentation~\cite{liu2019structured}, model compression~\cite{Li_2020_CVPR}, and improving 
the generalization capacity of DNNs~\cite{furlanello2018born}.

Although knowledge distillation has achieved success in various applications, few works have attempted to explain how and why this approach helps the training of the DNN. Hinton~\textit{et al.}~\cite{hinton2015distilling} demonstrated that learning from ``soft targets'' was easier than learning from ``hard targets".
Furlanello~\textit{et al.}~\cite{furlanello2018born} explained the efficiency of knowledge distillation as re-weighting training examples.
From a theoretical perspective, Lopez-Paz~\textit{et al.}~\cite{lopez2015unifying} related knowledge distillation to learning from privileged information under a noise-free setting.  Phuong and Lampert~\cite{phuong2019towards} explained the success of knowledge distillation from the perspective of data distribution, optimization bias, and the size of the training set. 
Yuan~\textit{et al.} ~\cite{yuan2020revisiting} interpreted knowledge distillation in terms of label smoothing regularization. 
Menon~\textit{et al.}~\cite{menon2020distillation} considered that a good teacher modeling the Bayes class-probabilities resulted in the success of knowledge distillation.

However, previous explanations for knowledge distillation were mainly based on 
qualitative statements~\cite{furlanello2018born} or theoretically relied on strong assumptions such as linear models~\cite{phuong2019towards}.
In contrast, we interpret knowledge distillation from the perspective of knowledge representations, \textit{i.e.} quantifying, analyzing, and comparing knowledge points encoded in distilled DNNs and knowledge points encoded in DNNs learning from scratch.
Hence, knowledge points provide new insight into the success of knowledge distillation.

\section{Algorithm}
\label{sec:Algorithm}
In this study, we explain the success of knowledge distillation from a new perspective, \textit{i.e.} quantifying knowledge points encoded in the intermediate layers of a DNN for classification.
In contrast to previous explanations of knowledge distillation, our method enables people to explain the representation power of a DNN using the quantity and the quality of knowledge points encoded in a DNN, which reflects a more direct and stronger connection between the knowledge and performance.
To this end, we measure the information in each input unit discarded by the DNN to define knowledge points, based on the information theory.
We propose three hypotheses on the mechanism of knowledge distillation to explain its success.
To verify these hypotheses, we further design three types of metrics based on the quantification of knowledge points, which evaluate the representation quality of the DNN.
Note that theoretically, knowledge distillation is generally considered to involve two fully different utilities.
Distillation from high-dimensional intermediate-layer features usually exhibits the first type of utility,~\textit{i.e.,} forcing the student network to mimic knowledge points of the teacher network.
In contrast, distilling from relatively low-dimensional network output often exhibits the second type of utility,~\textit{i.e.,} selecting confident samples for learning and ignoring  unconfident samples.
In this paper, we mainly explain the first utility of knowledge distillation, which is mainly exhibited by distilling from high-dimensional intermediate-layer features.

\subsection{Quantifying Information Discarding}
Before defining and quantifying knowledge points encoded in DNNs, we first introduce how to measure the pixel-wise information discarding of the input sample in this section.

According to information-bottleneck theory~\cite{wolchover2017new,shwartz2017opening}, we consider the signal processing of a forward propagation as the layer-wise discarding of input information.
In low layers, most input information is used to compute features.
Whereas, in high layers, information in input units unrelated to the inference is gradually discarded, and only related pixel-wise information is maintained in the computation of network features.
In this way, we consider that features of the highest layer mainly encode the information, which is highly relevant to the inferences.

Thus, we propose methods to quantify the input information encoded in a specific intermediate layer of a DNN~\cite{guan2019towards,deepInfo}, \textit{i.e.} measuring how much input information is discarded when the DNN extracts the feature of this layer.
To this end, the information in each input unit discarded by the DNN is formulated as the entropy, given a feature of a certain network layer.

Specifically, given a trained DNN for classification and an object instance $x\in{R^{n}}$ with $n$ input units, let $f^{*}=f(x)$ denote the feature of a network intermediate layer, 
and let $y^{*}$ denote the network prediction~\textit{w.r.t.} $x$.
The input unit is referred to as a variable (or a set of variables) in the input sample.
For example, each pixel (or pixels within a small local region) of an input image can be taken as an input unit, 
and the embedding of each word in an input sentence also can be regarded as an input unit.

In our previous works~\cite{guan2019towards,deepInfo}, 
we consider that the DNN satisfies the Lipschitz constraint
$\parallel{y'-y^{*}}\parallel \le \kappa \parallel{f(x')-f^{*}}\parallel$, where $\kappa$ is the Lipschitz constant, and $y'$ corresponds to the network prediction of the sample $x'$.
The  Lipschitz constant $\kappa$ can be computed as the maximum norm of the gradient within a small range of features,~\textit{i.e.,}
{\small$\kappa = \sup_{f(x'):\parallel{f(x')-f^{*}}\parallel ^{2}\le \tau} \Vert \frac{\partial y'}{\partial f(x')} \Vert$}.
This indicates that 
if we weakly perturb the low-dimensional manifold of features \textit{w.r.t.} the input $x$ within a small range $\small{\{f(x')\big{|} \parallel{f(x')-f^{*}}\parallel ^{2}\le \tau\}}$, then the output of the DNN was also perturbed within a small range $\small{\parallel{y'-y^{*}}\parallel \le \kappa \cdot \tau}$, where $\tau$ is a  positive scalar.
In other words, all such weakly perturbed features usually represent the same object instance.

However, determining the explicit low-dimensional manifold of features \textit{w.r.t.} the input $x$ is very difficult.
As an expediency, we add perturbations $\Delta x$ to the input sample $x$ to approximate the manifold of the feature, \textit{i.e.} generating samples 
$x'= x+\Delta x$, subject to
$\small{ \parallel{f(x')-f^{*}}\parallel ^{2}\le \tau}$.
In this way, the entropy $H(X')$ measures the uncertainty of the input when the feature represents the same object instance, 
$\small{\parallel{y'-y^{*}}\parallel \le \kappa \cdot \tau}$, as mentioned above.
In other words, $H(X')$ quantifies how much input information can be discarded without affecting the inference of the object.

\begin{small}\begin{equation}\label{1}
H(X')   ~~~\textit{s.t.}  ~~\forall{x'\in{X'},} ~~\parallel{f(x')-f^{*}}\parallel ^{2} \le \tau.
\end{equation}\end{small}
$X'$ denotes a set of inputs corresponding to the concept of a specific object instance. 
We assume that $\Delta x$ is an \textit{i.i.d.} Gaussian noise, thereby $x'= x+\Delta x \sim \mathcal{N}(x,\Sigma(\boldsymbol\sigma) = diag(\sigma_{1}^{2},\dots, \sigma_{n}^{2}))$.
Here, $\sigma_{i}$ indicates the variance of the perturbation \textit{w.r.t.} the $i$-th input unit.
The assumption of the Gaussian distribution ensures that the entropy $H(X') $ of the input can be decomposed into pixel-level entropies $\{H_{i}\}$.

\begin{small}
\begin{equation}\label{2}
H(X') = \sum_{i=1}^{n} H_{i},
\end{equation}
\end{small}
where $H_{i} = \log\sigma_{i} +  \frac{1}{2} \log(2\pi{e})$ measures the pixel-wise information discarding. 
A large value of $H_{i}$ usually indicates that the information of the $i$-th unit is significantly discarded during the forward propagation.
For example, as shown in Figure~\ref{fig:layer}(a), information encoded in background pixels is significantly discarded in the fine-grained bird classification.

Thus, we learn the optimal $\Sigma(\boldsymbol\sigma) = diag(\sigma_{1}^{2},\dots, \sigma_{n}^{2})$ via the maximum-entropy principle to compute the pixel-wise information discarding $H_{i}= \log\sigma_{i} +  \frac{1}{2} \log(2\pi{e})$.
The objective function is formulated in Equation~\eqref{1}, which maximizes $H(X')$ subject to constraining features within the scope of a specific object instance $\small{\parallel{f(x')-f^{*}}\parallel ^{2}\le \tau}$.
That is, we enumerate all perturbation directions in the input sample $x$ within a small variance of the feature $f^{*}$, in order to approximate the low-dimensional manifold of $f(x')$. 
We use the Lagrange multiplier to approximate Equation~\eqref{1} as the following loss.

\begin{small}
\begin{align}
&\qquad\min_{{\boldsymbol\sigma}} \quad \text{Loss}({\boldsymbol\sigma}),\nonumber \\
\text{Loss}({\boldsymbol\sigma}) 
\!\!=&\frac{1}{\delta_f^2}\mathop{\mathbb{E}}\limits_{x' \sim \mathcal{N}(x,\Sigma(\boldsymbol\sigma))}\left[\Vert f(x')-f^{*}\Vert^2\right]-\alpha\sum_{i=1}^{n}H_i,\label{eqn:sigma loss pre} \\
=& \frac{1}{\delta_f^2}\mathop{\mathbb{E}}\limits_{x'=x+{\boldsymbol\sigma}\circ{\boldsymbol\delta},\atop{\boldsymbol\delta}\sim\mathcal{N}({\bf 0},I)}\left[\Vert f(x')-f^{*}\Vert^2\right]-\alpha\sum_{i=1}^{n}H_i.\label{eqn:sigma loss}
\end{align}
\end{small}
Considering that Equation~\eqref{eqn:sigma loss pre} is intractable, we use $x'=x+{\boldsymbol\sigma}\circ{\boldsymbol\delta}, \delta\sim\mathcal{N}(0,I)$ to simplify the computation of $\text{Loss}({\boldsymbol\sigma})$, where {\small$\circ$} denotes the element-wise multiplication.
{\small${\boldsymbol\sigma}=[\sigma_1,\ldots,\sigma_{n}]^{\top}$} represents the range that the input sample can change.
In this way, $\text{Loss}({\boldsymbol\sigma})$ in Equation~\eqref{eqn:sigma loss} is tractable, and we can learn $\boldsymbol\sigma$ via gradient descent.
{\small $\delta_f^2 = \lim_{\tau\rightarrow0^{+}}\mathbb{E}_{x'\sim\mathcal{N}(x,\tau{^2} I )}\left[\Vert f(x')-f^{*}\Vert^2\right]$} denotes the inherent variance of intermediate-layer features subject to small input noises with a fixed magnitude $\tau$. $\delta_f^2$ is used for normalization, and a positive scalar $\alpha$ is used to balance the two loss terms.

\subsection{Quantification of knowledge points}
\label{hypothesis1}
\begin{framed}
\label{Quantification of knowledge points}
\textbf{Hypothesis 1}: Knowledge distillation makes the DNN encode more knowledge points than learning from scratch.
\end{framed}

The basic idea is to measure the quantity and quality of knowledge points encoded in the intermediate layers of a DNN.

\begin{figure*}[t]
    \centering
    \includegraphics[width=0.83\linewidth]{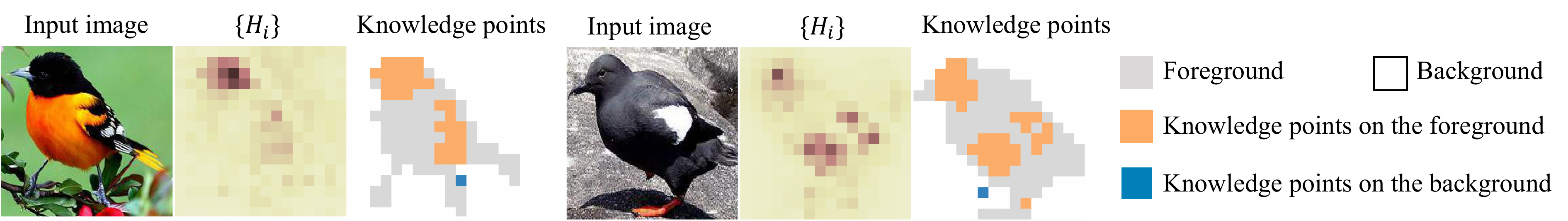}
    \caption{Visualization of knowledge points. The dark color indicates the low-entropy value ${H_{i}}$. Image regions with low pixel-wise entropies ${\{H_{i}\}}$ are considered as knowledge points. In a real implementation, to improve the stability and efficiency of the computation, we divide the image into $16\times16$ grids, and each grid is taken as a ``pixel" to compute the entropy $H_{i}$.}
    \label{fig1}
    \vspace{-5pt}
\end{figure*}

$\bullet\quad$\textbf{The quantity of knowledge points.}
In this paper, we use the pixel-wise information of each input unit discarded by the DNN to define and quantify knowledge points.
Given a trained DNN and an input sample $x \in X$, let us consider the pixel-wise information discarding ${H_{i}}$ \textit{w.r.t.} the feature of a certain network layer $f^{*}=f(x)$.
$X$ represents a set of input samples.

According to Equation~\eqref{2}, a low pixel-wise entropy $H_{i}$ represents that the information of this input unit is less discarded.
In other words, the DNN tends to use these input units with low entropies to compute the feature $f^{*}$ and make inferences.
Thus, a knowledge point is defined as an input unit with a low pixel-wise entropy $H_{i}$, which encodes discriminative information for prediction.
For example, the heads of birs in Figure~\ref{fig1} are referred to as knowledge points, which are useful for the fine-grained bird classification.

To this end, we use the average entropy $\overline{H}$ of all background units $\Lambda_{\textrm{bg}}$ as a baseline to determine knowledge points, \textit{i.e.} $\overline{H}=\mathbb{E}_{i \in \Lambda_{\textrm{bg}}}[H_{i}]$.
This is because information encoded in background units is usually supposed to be significantly discarded and irrelevant to the inference.
If the entropy of a unit is significantly lower than the baseline entropy $\overline{H} - H_{i} > b$, then this input unit can be considered as a valid knowledge point.
Here, $b$ is a threshold to determine the knowledge points, which is introduced in Section~\ref{Implementation Details}.
In this way, we can quantify the number of knowledge points in the DNN.

$\bullet\quad$\textbf{The quality of knowledge points.}
In addition to the quantity of knowledge points, we also consider the quality of knowledge points.
Generally speaking, information encoded in foreground units is usually supposed to be crucial to the inferences.
In comparison, information encoded in background units is often supposed to exhibit negligible effects on the prediction.
Therefore, we design a metric $\lambda$ to evaluate the quality of knowledge points by examining whether or not most knowledge points are localized in the foreground.

\begin{small}
\begin{align}
&\lambda = \mathbb{E}_{x \in X}\Bigg[{N_{\textrm{point}}^{\textrm{fg}}(x)} /  \big( N_{\textrm{point}}^{\textrm{fg}}(x) + N_{\textrm{point}}^{\textrm{bg}}(x) \big) \Bigg],\\
&N_{\textrm{point}}^{\textrm{bg}}(x) =\sum_{i \in \Lambda_{\textrm{bg}}~\textrm{\textit{w.r.t.} }x} \mathbbm{1}(\overline{H} - H_{i} > b),\label{eqn:bg}\\
&N_{\textrm{point}}^{\textrm{fg}}(x) =\sum_{i \in \Lambda_{\textrm{fg}}~\textrm{\textit{w.r.t.} }x} \mathbbm{1}(\overline{H} - H_{i} > b).\label{eqn:fg}
\end{align}
\end{small}
$N_{\textrm{point}}^{\textrm{bg}}(x)$ and $N_{\textrm{point}}^{\textrm{fg}}(x)$ denote the number of knowledge points encoded in the background and the foreground, respectively.
$\Lambda_{\textrm{bg}}$ and $\Lambda_{\textrm{fg}}$ are sets of input units in the background and the foreground \textit{w.r.t.} the input sample $x$, respectively.
$\mathbbm{1}(\cdot)$ refers to the indicator function. If the condition inside is valid, then $\mathbbm{1}(\cdot)$ returns $1$; otherwise, returns $0$.
A large value of $\lambda$ represents that the feature representation of the DNN is reliable; otherwise, unreliable.

Thus, knowledge points provide a new way to explain the mechanism of knowledge distillation, \textit{i.e.} checking whether knowledge distillation makes the DNN encode a large amount of knowledge points, and whether most knowledge points are precisely localized in the foreground.


\begin{table}[t]
\begin{center}
\caption{Comparisons of different explanation methods in terms of the coherency. The coherency of the entropy-based method enables fair layer-wise comparisons within a network and between different networks.}
\label{Generality and coherency}
\resizebox{0.98\linewidth}{!}{\
\begin{tabular}{c| c| c}
\toprule
\multirow{4}*{} & \multicolumn{2}{c}{Coherency} \\
\cline{2-3}
&\!\! Fair layer-wise\!\!&\!\!\!Fair comparisons between\\
&\!\!comparisons\!\!\!\!&\!\!\!\!different networks \\
\midrule
Gradient-based~\cite{zhou2016learning,simonyan2017deep,yosinski2015understanding,mahendran2015understanding}\!\! &  No& No\\
\midrule
Perturbation-based~\cite{fong2017interpretable,kindermans2017learning}\!\! & No& No\\
\midrule
Inversion-based~\cite{dosovitskiy2016inverting}\!\!  &  No& No\\
\midrule
Entropy-based & Yes&  Yes\\
\bottomrule
\end{tabular}}
\end{center}
\vspace{-20pt}
\end{table}

\textbf{Coherency and generality}.
As discussed in \cite{guan2019towards,deepInfo}, a trustworthy explanation method to evaluate feature representations of the DNN should satisfy the criteria of the coherency and generality.
From this perspective, we analyze the trustworthiness of knowledge points.

\textbf{Coherency} indicates that an explanation method should reflect the essence of the encoded knowledge representations in DNNs, which are invariant to network settings.
In this study, we use pixel-wise information discarding to define knowledge points without strong assumptions on feature representations, or network architectures.
In this way, knowledge points provide a coherent evaluation of feature representations.
However, most existing explanation methods shown in Table~\ref{Generality and coherency} usually fail to meet this criterion due to their biased assumptions. 
For example, the CAM~\cite{zhou2016learning} and gradient explanations~\cite{simonyan2017deep} do not generate coherent explanations, because the gradient map $\frac{\partial Loss}{\partial f}$ is usually not coherent through different layers.

To this end, we compare our method with CAM and gradient explanations.
Theoretically, we can easily construct two DNNs to represent exactly the same knowledge but with different magnitudes of gradients, as follows.
A VGG-16 model~\cite{simonyan2015very} was learned using the CUB200-2011 dataset~\cite{wah2011caltech} for fine-grained bird classification.
Given this pre-trained DNN, we slightly revised the magnitude of parameters in every pair of neighboring convolutional layers {\small$y=x\otimes w+b$} to examine the coherency.
For the {\small$L$}-th and {\small$(L+1)$}-th layers, parameters were revised as {\small$w^{(L)}\leftarrow 4w^{(L)}$}, {\small$w^{(L+1)}\leftarrow w^{(L+1)}/4$}, {\small$b^{(L)}\leftarrow 4b^{(L)}$}, {\small$b^{(L+1)}\leftarrow b^{(L+1)}/4$}.
Such revisions did not change knowledge representations or the network output, but did alter the gradient magnitude.

As Figure~\ref{fig:layer}(b) shows, magnitudes of the CAM and the gradient explanation methods are sensitive to the magnitude of parameters. In comparison, our method is not affected by the magnitude of parameters, which demonstrates the coherency. Thus, the coherency of our method enables fair layer-wise comparisons within a DNN, as well as fair comparisons between different DNNs.

\textbf{Generality} means that an explanation method should have strong connections to existing mathematical theories.
To this end, we quantify knowledge points using the entropy of the input sample.
The entropy is a generic tool with strong connections to various theories, \textit{e.g.} the information-bottleneck theory~\cite{wolchover2017new,shwartz2017opening}.
As a result, the generality of knowledge points enables us to fairly compare feature representations between different DNNs.
Note that calculating knowledge points requires the annotation of the foreground, but this requirement does not hurt the coherency and the generality of knowledge points.

\subsection{Learning simultaneously or sequentially}
\label{hypothesis 2}
\begin{framed}
\textbf{Hypothesis 2}: Knowledge distillation makes the DNN more likely to learn different knowledge points simultaneously. Whereas, the DNN learning from scratch tends to encode various knowledge points sequentially in different epochs.
\end{framed}

\begin{figure}[t]
    \centering
    \includegraphics[width=0.7\linewidth]{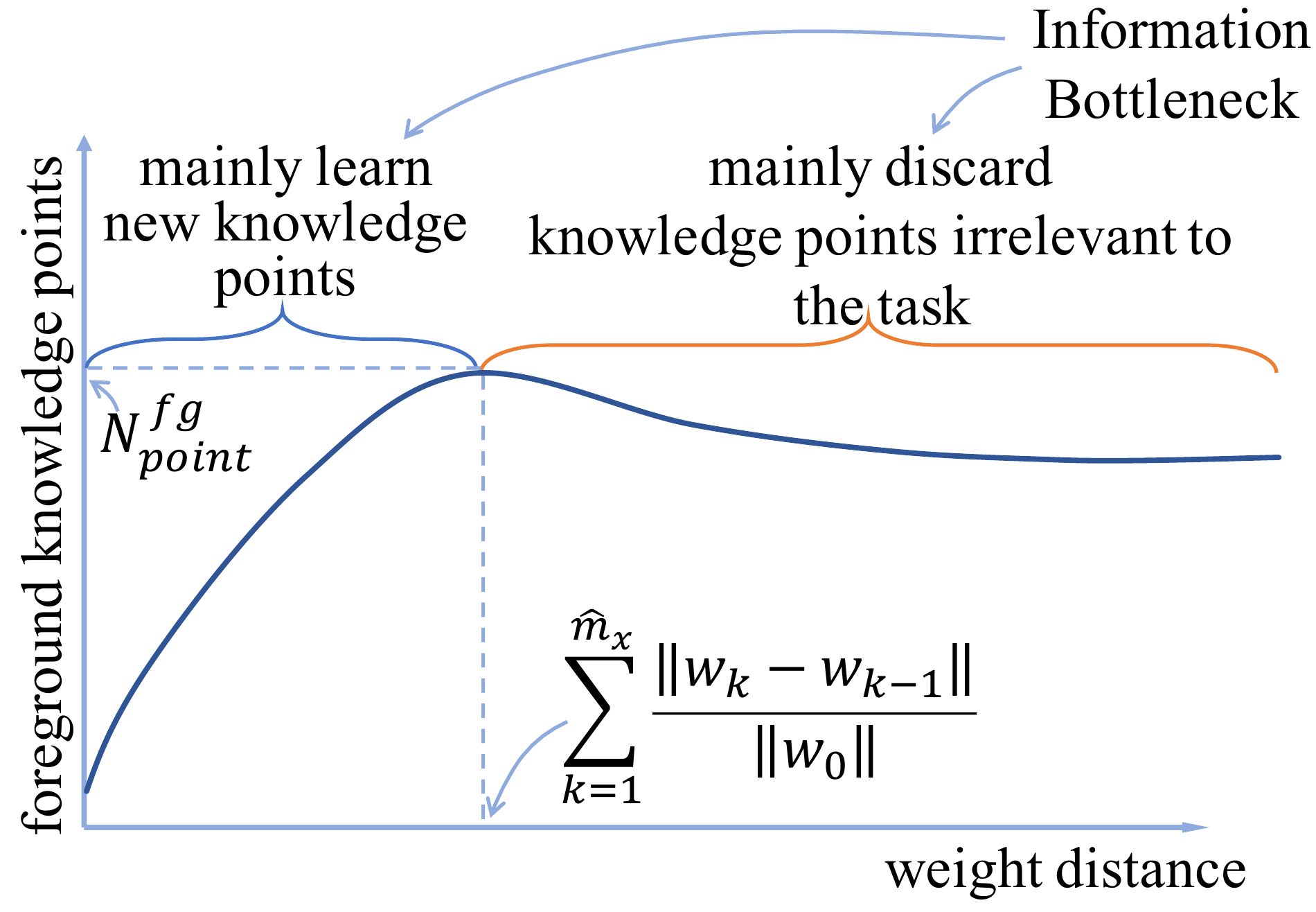}
	\vspace{-5pt}
	\caption{Procedure of learning foreground knowledge points \textit{w.r.t.} weight distances. According to information-bottleneck theory, a DNN tends to learn various knowledge points in early epochs and then discard knowledge points irrelevant to the task later. Strictly speaking, a DNN learns new knowledge points and discards old points during the entire course of the learning process. The epoch $\hat{m}_{x}$ encodes the richest foreground knowledge points.}
	\label{fig2}
	\vspace{-15pt}
\end{figure}

This hypothesis can be understood as follows.
Knowledge distillation forces the student network to mimic all knowledge points of the teacher network simultaneously.
In comparison, the DNN learning from scratch is more likely to learn different knowledge points in different epochs.
A DNN learning from scratch may encode knowledge points of common, simple, and elementary features in early epochs.
For example, in the scenario of fine-grained bird classification, the sky is a typical and simple feature to learn.
Whereas, knowledge points representing complex features may be encoded in later epochs by the DNN learning from scratch.
For example, in the scenario of fine-grained bird classification, the subtle differences between two similar bird species are difficult to learn.
In this way, the DNN learning from scratch tends to encode different knowledge points in a sequential manner, rather than simultaneously.

\textbf{Metrics.}
To verify Hypothesis 2, we design two metrics, which examines whether or not 
different knowledge points are learned within similar epochs, \textit{i.e.} whether different knowledge points are learned simultaneously.

Specifically, given a set of training samples $X$, let $M$ denote the total epochs used to train a DNN.
For each specific sample $x\in X$, foreground knowledge points learned by DNNs after $j$ epochs are represented as 
$N_{j}^{\textrm{fg}}(x)$, where $j \in \{1,2,\cdots,M\}$.

In this way, whether a DNN learns different knowledge points simultaneously can be measured from the following two aspects.
(1) Whether the number of foreground knowledge points $N_{j}^{\textrm{fg}}(x)$ increases fast along with the epoch number.
(2) Whether $N_{j}^{\textrm{fg}}(x)$~\textit{w.r.t.} different samples increase simultaneously.
In other words, the first aspect measures whether the DNN learns various knowledge points of a specific input sample quickly, and the second aspect evaluates whether a DNN learns knowledge points of different samples at similar speeds.

To this end, as shown in Figure \ref{fig2}, we consider the epoch $\hat{m}$ in which a DNN obtains the richest foreground knowledge points to measure the learning speeds, 
\textit{i.e.} $\hat{m}_{x} = \arg\max_{k} N_{k}^{\textrm{fg}}(x)$.
We use metrics $D_{\textrm{mean}}$ and $D_{\textrm{var}}$ based on the \textit{``weight distance'' } $\sum_{k=1}^{{m}} \frac{\|{w_{k}-w_{k-1}}\|}{\|{w_{0}}\|}$ to check whether a DNN learns knowledge points simultaneously.
$w_{0}$ and $w_{k}$ denote initial parameters and parameters learned after the $k$-th epoch, respectively.
The metric $D_{\textrm{mean}}$ represents the average weight distance at which the DNN obtains the richest foreground knowledge points, and
the metric $D_{\textrm{var}}$ describes the variation of the weight distance \textit{w.r.t.} different input samples.
These two metrics are defined as follows.

\begin{small}
\begin{equation}
\label{4}
\begin{split}
D_{\textrm{mean}} =\underset{x\in X}{\mathbb{E}}\bigg[ \sum\nolimits_{k=1}^{\hat{m}_{x}} \frac{\|{w_{k}-w_{k-1}}\|}{\|{w_{0}}\|} \bigg],\\
D_{\textrm{var}} = \underset{x\in X}{Var}\bigg[ \sum\nolimits_{k=1}^{\hat{m}_{x}} \frac{\|{w_{k}-w_{k-1}}\|}{\|{w_{0}}\|} \bigg].
\end{split}
\end{equation}
\end{small}
Here, we use the weight distance  to evaluate the learning effect at each epoch~\cite{garipov2018loss,flennerhag2018transferring}, rather than use the epoch number.
This is because the weight distance quantifies the total path of updating the parameter $w_{k}$ at each epoch $k$ better than the epoch number.
In other words, the weight distance can measure the dynamic learning process of a DNN better.
In this way, $D_{\textrm{mean}}$ measures whether a DNN learns knowledge points quickly, and $D_{\textrm{var}}$ evaluates whether a DNN learns knowledge points of different input samples at similar speeds.
Thus, small values of $D_{\textrm{mean}}$ and $D_{\textrm{std}}$ reflect that the DNN learns various knowledge points quickly and simultaneously.

\subsection {Learning with Less Detours}
\begin{framed}
\textbf{Hypothesis 3}: Knowledge distillation usually yields more stable optimization directions than learning from scratch.
\end{framed}

\begin{figure}[t]
    \centering
    \includegraphics[width=0.8\linewidth]{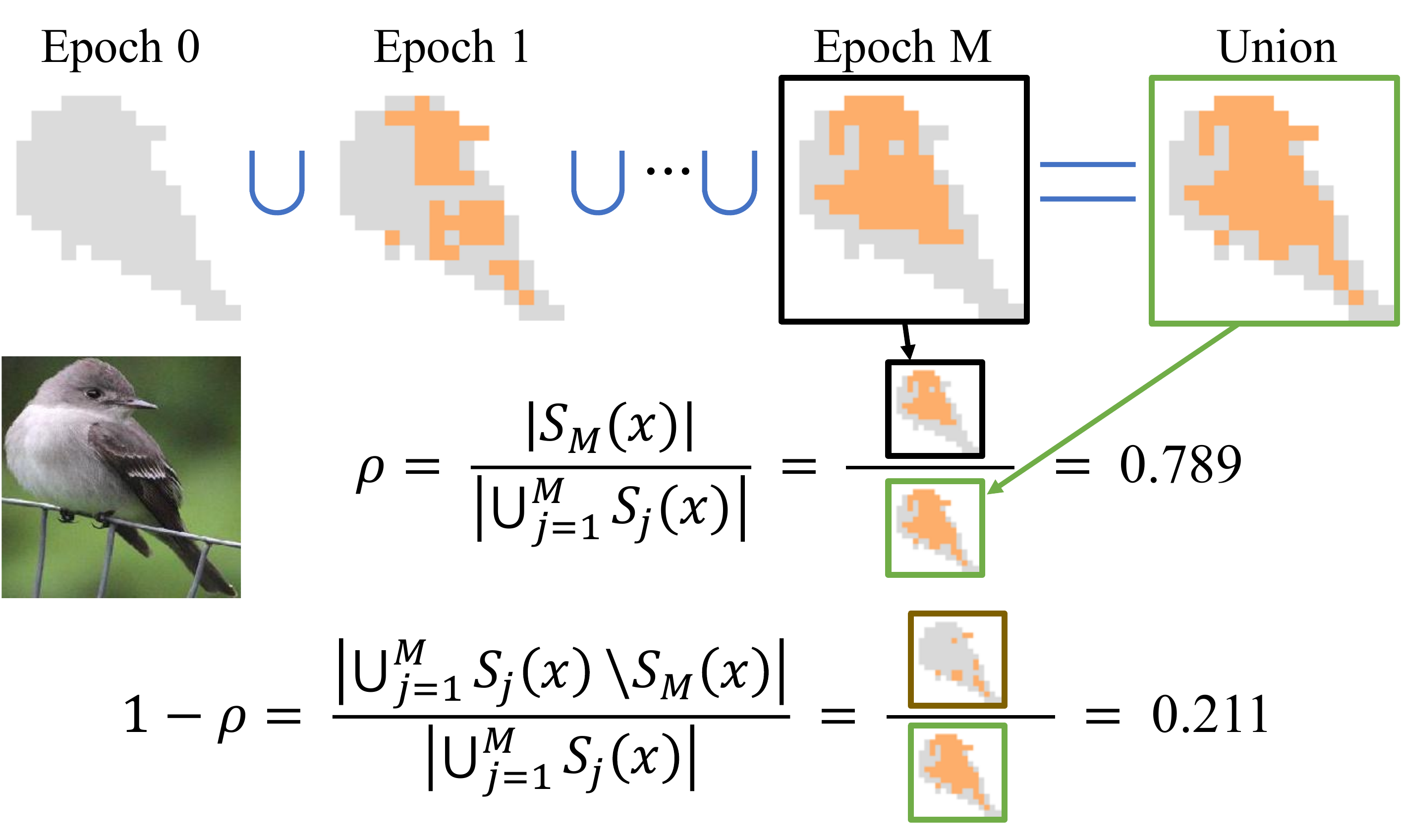}
    \caption{Detours of learning knowledge points. We visualize sets of foreground knowledge points learned after different epochs. The green box indicates the union of knowledge points learned during all epochs. The $(1-\rho)$ value denotes the ratio of knowledge points that are discarded during the learning process to the union set of all learned knowledge points. A larger $\rho$ value indicates DNN is optimized with fewer detours.}
    \vspace{-10pt}
   \label{fig3}
\end{figure}

This hypothesis can be understood as follows.
According to the information-bottleneck theory~\cite{wolchover2017new,shwartz2017opening}, the DNN learning from scratch usually tries to model various knowledge points in early epochs and then discards those unrelated to the inference in late epochs.
This means that the DNN learning from scratch cannot directly encode target knowledge points for inference.
Thus, the optimization direction of this DNN is inconsistent and unstable in early and late epochs, \textit{i.e.} with significant detours.
In contrast to learning from scratch, knowledge distillation forces the student network to directly mimic the well-trained teacher network.
With the guidance of the teacher network, knowledge distillation makes the student network straightforwardly model target knowledge points without temporarily modeling and discarding other knowledge points.
Thus, the DNN learning via knowledge distillation tends to be optimized in a stable direction, \textit{i.e.} without significant detours.

\textbf{Metric.}
To verify Hypothesis 3, we design a metric to evaluate whether a DNN is optimized in a stable and consistent direction.
Let $S_{j}(x)=\{i\in x|\overline{H} - H_{i} > b\}$ denote the set of foreground knowledge points in the input sample $I$ encoded by the DNN learned after the $j$-th epoch, where $j\in\{1,2,\cdots,M\}$.
Each knowledge point $a \in{S_{j}(x)}$ is referred to as a specific unit $i$ on the foreground of the input sample $I$, which satisfies $\overline{H} - H_{i} > b$. 
Then, we use the metric $\rho$ to measure the stability of optimization directions.
\begin{equation}\label{5}
\small{ \rho = \frac{|S_{M}(x)|}{|\bigcup_{j=1}^{M} S_{j}(x)|}.}
\end{equation}
$\rho$ represents the ratio of finally encoded knowledge points to all temporarily attempted knowledge points in intermediate epochs, where $|\cdot|$ denotes the cardinality of the set.
More specifically, the numerator of Equation~\eqref{5} reflects the number of foreground knowledge points, which have been chosen ultimately for inference.
For example, knowledge points shown in the black box of Figure \ref{fig3} are encoded as final knowledge points for fine-grained bird classification.
The denominator represents the number of all knowledge points temporarily learned during the entire training process, which is shown as the green box in Figure \ref{fig3}.
Moreover, $(\bigcup_{j=1}^{M} S_{j}(x)\setminus{S_{M}(x))}$ indicates a set of knowledge points, which have been tried, but finally are discarded by the DNN. 
The set of those discarded knowledge points is shown as the brown box in Figure \ref{fig3}.
Thus, a high value of $\rho$ indicates that the DNN is optimized without significant detours and more stably, and vice versa.

\section{Experiment}
In this section, we conducted experiments on image classification, 3D point cloud classification, and natural language processing (NLP) tasks. 
Experimental results verified the proposed hypotheses.

\subsection{Implementation Details}
\label{Implementation Details}

Given a teacher network, we distilled knowledge from the teacher network to the student network.
In order to remove the effects of network architectures, the student network had the same architecture as the teacher network.
For fair comparisons, the DNN learning from scratch was also required to have the same architecture as the teacher network.
For convenience,  we named the DNN learning from scratch the \textbf{baseline network} hereinafter. 

Additionally, we conducted experiments to check whether the size of the teacher network affected the verification of the three hypotheses.
Table~\ref{tab:teacher arch} and Table~\ref{tab:tiny} show that when the student network distilled from a larger teacher network, Hypotheses 1-3 were still verified.
This phenomenon indicates that variations in the size of the teacher network did not hurt the verification of our conclusions.
To simplify the story, we set student networks to have the same architectures as teacher networks in the following experiments.

\begin{table}[t]
\begin{center}
\caption{
Comparisons between the student network and the baseline network (B) for the image classification on the CUB200-2011 dataset, where the student network distilled from the $\textrm{FC}_{1}$ layer of a larger teacher network.}
\label{tab:teacher arch}
\resizebox{0.98\linewidth}{!}{\
\begin{tabular}{p{0.15\linewidth}<{\centering}| p{0.35\linewidth}<{\centering} |p{0.1\linewidth}<{\centering} p{0.1\linewidth}<{\centering}p{0.05\linewidth}<{\centering}| p{0.1\linewidth}<{\centering} p{0.1\linewidth}<{\centering}|p{0.05\linewidth}<{\centering}}
\toprule
\small{Target network}&\small{\makecell[c]{Learning\\ methods}}& $N_{\textrm{point}}^{\textrm{fg}}\uparrow$& $N_{\textrm{point}}^{\textrm{bg}}\downarrow$ & $\lambda\uparrow$ &$D_{\textrm{mean}}\downarrow$& $D_{\textrm{var}}\downarrow$&$\rho\uparrow$\\
\midrule

 \multirow{6}*{VGG-11}
 &\footnotesize{Distilling from VGG-11} &  {\textbf{34.06}}  & {\textbf{8.89}}&  {\textbf{0.80}}& \textbf{1.30}& \textbf{0.80}& \textbf{0.58}\\
& \footnotesize{Learning from scratch}  & 25.50  & 11.50 &0.70 &2.14 &3.29 &0.56 \\
\cline{2-8}
 
&  \footnotesize{Distilling from VGG-16} &  {\textbf{37.11}}  & 11.50&  {\textbf{0.78}}& \textbf{0.48}& \textbf{0.06}& \textbf{0.58}\\
& \footnotesize{Learning from scratch} &  25.50  & 11.50 &0.70 &2.14 &3.29 &0.56 \\
\cline{2-8}

&  \footnotesize{Distilling from VGG-19} & {\textbf{41.44}}  & \textbf{8.67}&  \textbf{0.83}& \textbf{0.55}& \textbf{0.10}& \textbf{0.63}\\
& \footnotesize{Learning from scratch} &  25.50  & 11.50 &0.70 &2.14 &3.29 &0.56 \\
\midrule

 \multirow{4}*{AlexNet}
&\scriptsize{Distilling from AlexNet} &  {\textbf{35.29}}  & {\textbf{4.07}}&  {\textbf{0.90}}& 4.54 &8.43 & \textbf{0.57}\\
& \footnotesize{Learning from scratch} & 24.00 & 5.90 &0.80 &\textbf{2.80} & \textbf{5.32} &0.53 \\
\cline{2-8}
 
 &  \footnotesize{Distilling from VGG-16} &  {\textbf{39.14}}  & \textbf{4.45}&  {\textbf{0.90}}& \textbf{0.78}& \textbf{0.05}& \textbf{0.67}\\
& \footnotesize{Learning from scratch} & 24.00 & 5.90 &0.80 &{2.80} & {5.32} &0.53 \\
\cline{2-8}

& \footnotesize{Distilling from VGG-19} & {\textbf{38.36}}  & \textbf{4.14}&  \textbf{0.90}& \textbf{0.92}& \textbf{0.11}& \textbf{0.65}\\
& \footnotesize{Learning from scratch} & 24.00 & 5.90 &0.80 &2.80 & 5.32 &0.53 \\

\bottomrule
\end{tabular}}
\end{center}
\vspace{-25pt}
\end{table}

\begin{table*}[t]
\begin{center}
\caption{Comparisons between the student network (S) and the baseline network (B) for image classification on the CUB200-2011 dataset, respectively. 
Metrics were evaluated at the $\textrm{FC}_{1}$ layer under different settings of the grid size and different settings of the threshold $b$. Experimental results verified Hypotheses 1-3.}
\label{tab:grid_size_b}
\resizebox{0.98\linewidth}{!}{\
\begin{tabular}{ p{0.1\linewidth}<{\centering} p{0.025\linewidth}<{\centering}|p{0.05\linewidth}<{\centering} p{0.05\linewidth}<{\centering}p{0.05\linewidth}<{\centering}| p{0.05\linewidth}<{\centering} p{0.05\linewidth}<{\centering}|p{0.05\linewidth}<{\centering}|
 p{0.1\linewidth}<{\centering} p{0.025\linewidth}<{\centering}|p{0.05\linewidth}<{\centering} p{0.05\linewidth}<{\centering}p{0.05\linewidth}<{\centering}| p{0.05\linewidth}<{\centering} p{0.05\linewidth}<{\centering}|p{0.05\linewidth}<{\centering}
}
\toprule
\multicolumn{8}{c|}{Based on VGG-16}&\multicolumn{8}{c}{Based on VGG-19}\\
\midrule

Grid size&&$N_{\textrm{point}}^{\textrm{fg}}\uparrow$& $N_{\textrm{point}}^{\textrm{bg}}\downarrow$ & $\lambda\uparrow$ &$D_{\textrm{mean}}\downarrow$& $D_{\textrm{var}}\downarrow$&$\rho\uparrow$&
Grid size&&$N_{\textrm{point}}^{\textrm{fg}}\uparrow$& $N_{\textrm{point}}^{\textrm{bg}}\downarrow$ & $\lambda\uparrow$ &$D_{\textrm{mean}}\downarrow$& $D_{\textrm{var}}\downarrow$&$\rho\uparrow$\\
\midrule

\multirow{2}*{$8 \times 8$}
&{S} &  {\textbf{10.06}}  &  {\textbf{2.56}}& {\textbf{0.80}}& \textbf{0.51}& \textbf{0.07}& \textbf{0.63}&
\multirow{2}*{$8 \times 8$}
&{S} &  {\textbf{9.73}}  &  {\textbf{2.53}}& {\textbf{0.79}}& \textbf{0.54}& \textbf{0.12}& \textbf{0.61}
\\
& {B} & 6.47  & {3.35} &0.67 &2.63 & 3.61 &0.56&
& {B} & 6.43  & {3.13} &0.67 &2.64 &2.90 &0.52 \\
\midrule

\multirow{2}*{$16 \times 16$}
&{S} &  {\textbf{43.77}}  &  {\textbf{8.73}}& {\textbf{0.84}}& \textbf{0.63}& \textbf{0.06}& \textbf{0.66}&
\multirow{2}*{$16 \times 16$}
&{S} &  {\textbf{40.74}}  &  {\textbf{10.42}}& {\textbf{0.80}}& \textbf{0.67}& \textbf{0.15}& \textbf{0.60}
\\
&{B} & 22.50  & {11.27} &0.68 &2.39 & 4.98&0.50&
&{B} & 22.41  & {11.19} &0.67 &2.32 & 3.67 &0.47
\\
\midrule

\multirow{2}*{$28 \times 28$}
&{S} &  {\textbf{82.70}}  &  {\textbf{9.27}}& {\textbf{0.90}}& \textbf{0.62}& \textbf{0.08}& \textbf{0.58}&
\multirow{2}*{$28 \times 28$}
&{S} &  {\textbf{78.43}}  &  {\textbf{14.46}}& {\textbf{0.85}}& \textbf{0.65}& \textbf{0.12}& \textbf{0.52}
\\
 &{B} & 50.77  & 21.13 &0.72 &2.32 & 4.76 &0.48&
 &{B} & 49.51  & 22.14 &0.71 &2.10 & 3.90 &0.45 \\
\midrule\midrule

{\small{Threshold b}}&&$N_{\textrm{point}}^{\textrm{fg}}\uparrow$& $N_{\textrm{point}}^{\textrm{bg}}\downarrow$ & $\lambda\uparrow$ &$D_{\textrm{mean}}\downarrow$& $D_{\textrm{var}}\downarrow$&$\rho\uparrow$&
{\small{Threshold b}} &&$N_{\textrm{point}}^{\textrm{fg}}\uparrow$& $N_{\textrm{point}}^{\textrm{bg}}\downarrow$ & $\lambda\uparrow$ &$D_{\textrm{mean}}\downarrow$& $D_{\textrm{var}}\downarrow$&$\rho\uparrow$\\
\midrule

\multirow{2}*{0.15}
&{S} &  {\textbf{54.53}}  &  14.37 & {\textbf{0.80}}& \textbf{0.66}& \textbf{0.06}& \textbf{0.69}&
\multirow{2}*{0.15}
&{S} &  {\textbf{52.33}}  &  15.89 & {\textbf{0.77}}& \textbf{0.63}& \textbf{0.12}& \textbf{0.63}
\\
&{B} & 26.43  & \textbf{13.37} &0.68 &1.50 & 4.12 & 0.49&
&{B} & 26.04  & \textbf{13.22} &0.66 &1.45 &3.18 &0.44 \\
\midrule

\multirow{2}*{0.2}
&{S} &  {\textbf{43.77}}  &  {\textbf{8.73}}& {\textbf{0.84}}& \textbf{0.63}& \textbf{0.06}& \textbf{0.66}&
\multirow{2}*{0.2}
&{S} &  {\textbf{40.74}}  &  {\textbf{10.42}}& {\textbf{0.80}}& \textbf{0.67}& \textbf{0.15}& \textbf{0.60}
\\
 &{B} & 22.50  & {11.27} &0.68 &2.39 & 4.98&0.50&
&{B} & 22.41  & {11.19} &0.67 &2.32 & 3.67 &0.47
\\
\midrule

\multirow{2}*{0.25}
&{S} &  {\textbf{34.73}}  &  {\textbf{5.13}}& {\textbf{0.87}}& \textbf{0.61}& \textbf{0.07}& \textbf{0.62}&
\multirow{2}*{0.25}
&{S} &  {\textbf{32.23}}  &  {\textbf{6.97}}& {\textbf{0.83}}& \textbf{0.64}& \textbf{0.13}& \textbf{0.55}
\\
&{B} & 19.90  & 9.57 &0.68 &2.70 & 5.06 &0.53&
&{B} & 19.84  & 29.35 &0.68 &2.49 & 3.21 &0.48 \\

\bottomrule
\end{tabular}}
\end{center}
\vspace{-10pt}
\end{table*}


\textbf{Datasets \& DNNs.} 
\textit{For the image classification task}, we conducted experiments based on AlexNet~\cite{krizhevsky2012imagenet}, VGG-11, VGG-16, VGG-19~\cite{simonyan2015very}, ResNet-50, ResNet-101, and ResNet-152~\cite{he2016deep}. 
We trained these DNNs based on the ILSVRC-2013 DET dataset~\cite{sermanet2013overfeat}, the CUB200-2011 dataset~\cite{wah2011caltech}, and the Pascal VOC 2012 dataset~\cite{Everingham10} for object classification, respectively. 
Considering the high computational burden of training on the entire ILSVRC-2013 DET dataset, we conducted the classification of terrestrial mammal categories for comparative experiments.
Note that all teacher networks in Sections~\ref{Verification of Hypothesis 2},~\ref{Verification of Hypothesis 3} were pre-trained on the ImageNet dataset~\cite{russakovsky2015imagenet}, and then fine-tuned using the aforementioned dataset. 
In comparison, all baseline networks were trained from scratch. 
Moreover, data augmentation~\cite{jacobsen2018revnet} was applied to 
both the student network and the baseline network, when the DNNs were trained using the ILSVRC-2013 DET dataset or the Pascal VOC 2012 dataset.

We used object images cropped by object bounding boxes to train the aforementioned DNNs on each above dataset, respectively.
In particular, to achieve stable results, images in the Pascal VOC 2012 dataset were cropped by using $1.2 \,{width} \times 1.2 \,{height}$ of the original object bounding box. 
For the ILSVRC-2013 DET dataset, we cropped each image by using $1.5 \,{width} \times 1.5 \,{height}$ of the original object bounding box. 
This was performed because no ground-truth annotations of object segmentation were available for the ILSVRC-2013 DET dataset.
We used the object bounding box to separate foreground regions and background regions of images in the ILSVRC-2013 DET dataset.
In this way, pixels within the object bounding box were regarded as the foreground~$\Lambda_{\textrm{fg}}$, and pixels outside the object bounding box were referred to as the background~$\Lambda_{\textrm{bg}}$.

\textit{For natural language processing}, we fine-tuned a pre-trained BERT model~\cite{devlin2018bert} using the SQuAD dataset~\cite{rajpurkar-etal-2016-squad} as the teacher network towards the question-answering task.
Besides, we also fine-tuned another pre-trained BERT model on the SST-2 dataset~\cite{socher2013recursive} as the teacher network for binary sentiment classification.
Baseline networks and student networks were simply trained using samples in the SST-2 dataset or the SQuAD dataset.
Moreover, for both the SQuAD dataset and the SST-2 dataset, we annotated input units irrelevant to the prediction as the  background~$\Lambda_{\textrm{bg}}$, according to the human cognition.
We annotated units relevant to the inference as the foreground~$\Lambda_{\textrm{fg}}$.
For example, words related to answers were labeled as foregrounds in the SQuAD dataset. Words containing sentiments were annotated as foregrounds in the SST-2 dataset.

\textit{For 3D point cloud classification task}, we conducted comparative experiments on the PointNet~\cite{qi2017pointnet}, DGCNN~\cite{wang2019dynamic}, PointConv~\cite{wu2019pointconv}, and PointNet++~\cite{qi2017pointnet++} models.
Considering most widely used benchmark datasets for point cloud classification only contained foreground objects, we constructed a new dataset containing both foreground objects and background objects based on the ModelNet40 dataset~\cite{wu20153d}, as follows.
For each sample (\textit{i.e.} foreground object) in the ModelNet40 dataset, 
we first randomly sampled five point clouds, the labels of which differed from the foreground object. 
Then, we randomly selected $500$ points from the sampled five point clouds, and attached these points to the foreground object as background points.
Moreover, the teacher network was trained using all training data in this generated dataset for classification.
In comparison, we only randomly sampled $10\%$ training data in this generated dataset to learn baseline and student networks, respectively.
Thus, the teacher network was guaranteed to encode better knowledge representations.


\textbf{Distillation.} 
Given a well-trained teacher network and an input sample $x$, we selected a convolutional layer or a fully-connected (FC) layer $l$ as the target layer to perform knowledge distillation.
We used the distillation loss $\|f_{T}(x)-f_{S}(x)\|^{2}$ to force the student network to mimic the feature of the teacher network.
Here, $f_{T}(x)$ denoted the feature of $l$-th layer in the teacher network, and $f_{S}(x)$ indicated the feature of $l$-th layer in its corresponding student network.

In order to simplify the story, we used the distillation loss to learn the student network, rather than the classification loss.
Given the distilled features, we further trained parameters above the target layer $l$ merely using the classification loss.
In this way, we were able to ensure that all knowledge points of the student network were exclusively learned via distillation.
Otherwise, if we trained the student network using both the distillation loss and the classification loss, distinguishing whether the classification loss or the distillation loss was responsible for the learning of each knowledge point would be difficult.

Moreover, we employed two schemes of knowledge distillation in this study, including learning from intermediate-layer features and learning from network output.
Specifically, 
we distilled features in the top convolutional layer ($\textrm{Conv}$) or in each of three FC layers (namely $\textrm{FC}_{1}$, $\textrm{FC}_{2}$ and $\textrm{FC}_{3}$) of the teacher network to train the student network, respectively.
In this way, knowledge distillation from the $\textrm{FC}_{3}$ layer corresponded to the scheme of learning from output, and distillation from other layers corresponded to the scheme of learning from intermediate-layer features.
In spite of these two schemes, we mainly explained knowledge distillation from intermediate-layer features 
that forced the student network to mimic the knowledge of the teacher network as discussed above.

\textbf{Layers used to quantify knowledge points.} 
For each pair of the student and baseline networks, we measured knowledge points on the top convolutional layer and all FC layers.

Particularly, for DNNs trained on the ILSVRC-2013 DET dataset and the Pascal VOC 2012 dataset, we quantified knowledge points only on $\textrm{FC}_{1}$ and $\textrm{FC}_{2}$ layers,
because the dimension of features in the output layer $\textrm{FC}_{3}$ was much lower than that of intermediate-layer features in the $\textrm{FC}_{1}$ and $ \textrm{FC}_{2}$ layers.
Similarly, for DNNs trained on the ModelNet40 dataset, we also used the $\textrm{FC}_{1}$ and $ \textrm{FC}_{2}$ layers to quantify knowledge points.
For the BERT model, we measured knowledge points in its last hidden layer.
Considering ResNets usually only had a single FC layer, to obtain rich knowledge representations, we added two $\textrm{Conv}$ layers and two FC layers before the final FC layer.
In this way, we quantified knowledge points on these three FC layers.

\textbf{Hyper-parameter settings.}
We used the SGD optimizer, and set the batch size as 64 to train DNNs. 
We set the grid size to $16 \times 16$, and settings of the threshold $b$ were as follows.
For DNNs towards image classification, $b$ was set to $0.2$, except that we set $b$ to $0.25$ for AlexNet.
For BERT towards NLP tasks, $b$ was set to $0.2$.
For DNNs towards 3D point cloud classification,  $b$ was set to $0.25$, except that $b$ was set to $0.5$ for PointNet.

Additionally, we also tested the influence of the grid size and the threshold $b$ on the verification of the three hypotheses.
To this end, we conducted experiments with different settings of the grid size and different settings of the threshold $b$.
Table~\ref{tab:grid_size_b} shows that Hypotheses 1-3 were still validated, when knowledge points were calculated under different settings of the grid size and different settings of the threshold $b$.


\begin{table}[t]
\begin{center}
\caption{Comparisons of knowledge points encoded in the teacher network (T), the student network (S) and the baseline network (B) for image classification. The teacher network encoded more foreground knowledge points $N_{\textrm{point}}^{\textrm{fg}}$ and obtained a larger ratio $\lambda$ than the student network. Meanwhile, the student network had larger values of $N_{\textrm{point}}^{\textrm{fg}}$ and $\lambda$ than  baseline network.}
\label{teacher supplementary}
\resizebox{0.85\linewidth}{!}{\
\begin{tabular}{p{0.2\linewidth}<{\centering}| p{0.2\linewidth}<{\centering} p{0.1\linewidth}<{\centering}| p{0.2\linewidth}<{\centering} p{0.1\linewidth}<{\centering}}
\toprule
Dataset &Layer& &$N_{\textrm{point}}^{\textrm{fg}}\uparrow$&  $\lambda\uparrow$\\
\midrule
  \multirow{9}*{CUB}&\multirow{3}*{\footnotesize{VGG-16 $\textrm{FC}_{1}$}}&{T} & \textbf{34.00}  & \textbf{0.78}\\
& &{S} & {29.57}  & {0.75} \\
& &{B} & 22.50 & 0.68 \\
\cline{2-5}
 &\multirow{3}*{\footnotesize{VGG-16 $\textrm{FC}_{2}$}}&T & \textbf{34.62}  & \textbf{0.80} \\
& &S & {32.92}  & {0.75} \\
& &B & 23.31 & 0.67 \\
\cline{2-5}
&\multirow{3}*{\footnotesize{VGG-16 $\textrm{FC}_{3}$}}&T & \textbf{33.97}  & \textbf{0.81}\\
& &S & {29.78}  & 0.63 \\
& &B & {23.26} & {0.71} \\
\midrule

 \multirow{6}*{ILSVRC}&\multirow{3}*{\footnotesize{VGG-16 $\textrm{FC}_{1}$}}&T & \textbf{36.80} & \textbf{0.87} \\
& &S & {35.98} & {0.84} \\
& &B & {36.47}& 0.81 \\
\cline{2-5}
 &\multirow{3}*{\footnotesize{VGG-16 $\textrm{FC}_{2}$}}&T & 38.76 & \textbf{0.89} \\
& &S & \textbf{42.74} & {0.82}\\
& &B & {36.35} & 0.82 \\
\bottomrule
\end{tabular}}
\end{center}\vspace{-15pt}
\end{table}

\begin{figure*}[t]
    \centering
    \includegraphics[width=0.95\linewidth]{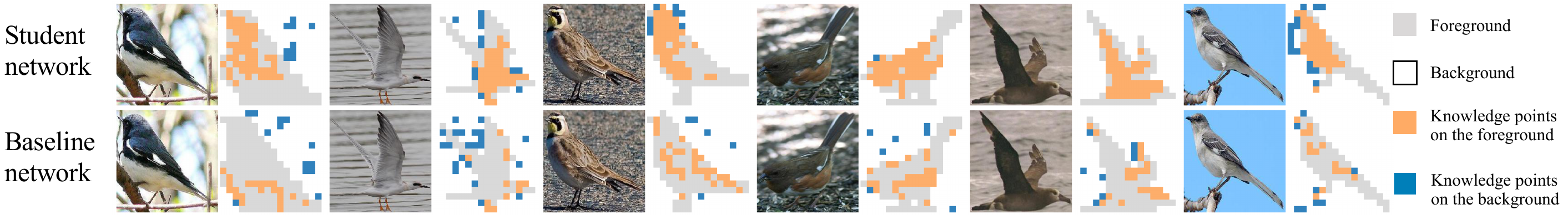}\vspace{-5pt}
    \caption{Visualization of knowledge points encoded in the $ \textrm{FC}_{1}$ layer of VGG-11. Generally, the student network exhibited a larger $N_{\textrm{point}}^{\textrm{fg}}$ value and a smaller $N_{\textrm{point}}^{\textrm{bg}}$ value than the baseline network.}
    \label{fig4}
    \vspace{-10pt}
\end{figure*}


\begin{table*}[t]
\begin{center}
\caption{Comparisons between the student network (S) and the baseline network (B) for image classification. In general, the student network exhibited larger values of $N_{\textrm{point}}^{\textrm{fg}}$, $\lambda$, $\rho$, and smaller values of $N_{\textrm{point}}^{\textrm{bg}}$, $D_{\textrm{mean}}$, $D_{\textrm{std}}$ than the baseline network, which verified Hypotheses 1-3.}
\label{tab:image class}
\resizebox{\linewidth}{!}{\
\settowidth\rotheadsize{\theadfont CUB200-2011 ILSVRC-2013 DET dataset Pascal VOC 2012 dataset}
\begin{tabular}{p{0.05\linewidth}<{\centering} p{0.01\linewidth}<{\centering} p{0.01\linewidth}<{\centering}|p{0.008\linewidth}<{\centering} |p{0.05\linewidth}<{\centering}p{0.05\linewidth}<{\centering} p{0.03\linewidth}<{\centering}| p{0.05\linewidth}<{\centering} p{0.06\linewidth}<{\centering}| p{0.03\linewidth}<{\centering}|p{0.008\linewidth}<{\centering} |p{0.05\linewidth}<{\centering}p{0.05\linewidth}<{\centering} p{0.03\linewidth}<{\centering}| p{0.05\linewidth}<{\centering} p{0.05\linewidth}<{\centering}| p{0.03\linewidth}<{\centering}|p{0.008\linewidth}<{\centering} |p{0.05\linewidth}<{\centering}p{0.05\linewidth}<{\centering} p{0.03\linewidth}<{\centering}| p{0.05\linewidth}<{\centering} p{0.05\linewidth}<{\centering}| p{0.03\linewidth}<{\centering}}
\toprule
\scriptsize{Network} &\scriptsize{Layer}& & &\scriptsize{$N_{\textrm{point}}^{\textrm{fg}}\uparrow$}& \scriptsize{$N_{\textrm{point}}^{\textrm{bg}}\downarrow$}&\scriptsize{$\lambda\uparrow$} &\scriptsize{$D_{\textrm{mean}}\downarrow$}& \scriptsize{$D_{\textrm{var}}\downarrow$}& \scriptsize{$\rho\uparrow$}& &\scriptsize{$N_{\textrm{point}}^{\textrm{fg}}\uparrow$}& \scriptsize{$N_{\textrm{point}}^{\textrm{bg}}\downarrow$}&\scriptsize{$\lambda\uparrow$} &\scriptsize{$D_{\textrm{mean}}\downarrow$}& \scriptsize{$D_{\textrm{var}}\downarrow$}& \scriptsize{$\rho\uparrow$}& &\scriptsize{$N_{\textrm{point}}^{\textrm{fg}}\uparrow$}& \scriptsize{$N_{\textrm{point}}^{\textrm{bg}}\downarrow$}&\scriptsize{$\lambda\uparrow$} &\scriptsize{$D_{\textrm{mean}}\downarrow$}& \scriptsize{$D_{\textrm{var}}\downarrow$}& \scriptsize{$\rho\uparrow$}\\
\cline{1-3} \cline{5-10} \cline{12-17} \cline{19-24}
 \multirow{8}*{\footnotesize{AlexNet}}
 &\multirow{2}*{\scriptsize{$\textrm{Conv}$}}&\footnotesize{S} &\multirow{36}*{\rothead{CUB200-2011 dataset}} & \textbf{26.59}&  \textbf{4.41}&\textbf{0.86} & \textbf{1.92}& \textbf{1.18}&0.49 &\multirow{36}*{\rothead{ILSVRC-2013 DET dataset}}&$-$&$-$ &$-$&$-$&$-$&$-$ &\multirow{36}*{\rothead{Pascal VOC 2012 dataset}}&\textbf{17.11} & \textbf{3.41} & \textbf{0.80} & \textbf{1.23}& \textbf{1.60}& 0.29\\
& &\footnotesize{B} & &24.54 &7.43 &0.77&4.13& 9.44&\textbf{0.57}& 
&$-$&$-$ &$-$&$-$&$-$&$-$ & 
& 16.24 & 4.78& 0.77 & 3.22& 2.64& \textbf{0.47}\\
 \cline{5-10} \cline{12-17} \cline{19-24}
 &\multirow{2}*{\scriptsize{$\textrm{FC}_{1}$}}&\footnotesize{S} & & \textbf{36.60}&  \textbf{4.00}&\textbf{0.90} & 5.13& 9.47& \textbf{0.57}&&\textbf{49.46} &\textbf{0.66} & \textbf{0.99} & \textbf{0.48}& \textbf{0.10}&\textbf{0.62} &&\textbf{25.84} & \textbf{5.86} & \textbf{0.79} & \textbf{1.14}& \textbf{0.56}& {0.43}\\
& &\footnotesize{B} & &24.13 &5.65 &0.81&\textbf{2.96}& \textbf{5.49}&0.52& &41.00 & 0.92&0.98 & 1.32& 0.31& 0.61& & 20.30 & 6.08&0.77 & 2.00& 2.21& \textbf{0.44}\\
\cline{5-10} \cline{12-17} \cline{19-24}
&\multirow{2}*{\scriptsize{$\textrm{FC}_{2}$}}&\footnotesize{S} & &\textbf{38.13}&  \textbf{3.50}&\textbf{0.92} &\textbf{3.77}& \textbf{3.97}& {0.49} & & \textbf{57.86}& {1.70}& {0.98}& \textbf{0.28}& \textbf{0.01}& {0.60}& & \textbf{31.81}&  {7.29}& \textbf{0.81}& \textbf{0.62}& \textbf{0.07}& \textbf{0.47}\\
& &\footnotesize{B}& &23.33 &5.48 &0.80&5.36& 20.79&\textbf{0.49} & & 42.24& \textbf{0.96} & \textbf{0.98} & 1.15& 0.15& \textbf{0.60}& & 21.85& \textbf{6.56} & 0.77 & 2.04& 1.46& 0.44\\
\cline{5-10} \cline{12-17} \cline{19-24}
&\multirow{2}*{\scriptsize{$\textrm{FC}_{3}$}}&\footnotesize{S} & & \textbf{33.20} &  \textbf{4.31}& \textbf{0.89} & \textbf{8.13} & \textbf{39.79} & \textbf{0.51}&&$-$&$-$ &$-$&$-$&$-$& $-$& &$-$  &$-$& $-$&$-$&$-$ &$-$\\
& &\footnotesize{B}& &22.73 & 4.94 & 0.83& 13.57 & 137.74 & 0.42& &$-$&$-$ &$-$&$-$&$-$&$-$ & & $-$ &$-$&$-$ &$-$&$-$&$-$\\
\cline{1-3} \cline{5-10} \cline{12-17} \cline{19-24}

 \multirow{8}*{\footnotesize{VGG-11}}
 &\multirow{2}*{\scriptsize{$\textrm{Conv}$}}&\footnotesize{S} & 
 & \textbf{34.54}&  \textbf{10.61}&\textbf{0.78} &1.76 & 2.24 &0.55  & 
&$-$&$-$ &$-$&$-$&$-$&$-$ &
 & \textbf{34.85}&  {10.59}&\textbf{0.75} &\textbf{1.32}&2.18 & \textbf{0.49}\\
& &\footnotesize{B} & 
& 30.11&  10.93& 0.73& \textbf{1.26} &  \textbf{1.19} &\textbf{0.63} & 
&$-$&$-$ &$-$&$-$&$-$&$-$ &
& 27.36& \textbf {8.85}&0.74 &1.33&\textbf {1.44} & 0.42\\
\cline{5-10} \cline{12-17} \cline{19-24}
 &\multirow{2}*{\scriptsize{$\textrm{FC}_{1}$}}&\footnotesize{S} & & \textbf{30.69}&  \textbf{10.65}&\textbf{0.75} &\textbf{1.21}& \textbf{0.61}& \textbf{0.56} & & \textbf{44.48}&  \textbf{4.68}&\textbf{0.91} &\textbf{0.26}& \textbf{0.06}& {0.50}& & \textbf{30.56}&  {8.36}&\textbf{0.78} &\textbf{1.09}& \textbf{0.30}& {0.38}\\
& &\footnotesize{B} & &24.26 & 10.77 &0.70 &2.01 & 3.18 &0.55& & 28.27& 7.80&0.80 & 0.93 & 0.08 & \textbf{0.53}& &20.31 & \textbf{7.28}& 0.73&  1.41&  0.54& \textbf{0.44}\\
\cline{5-10} \cline{12-17} \cline{19-24}
&\multirow{2}*{\scriptsize{$\textrm{FC}_{2}$}}&\footnotesize{S} & &\textbf{36.51}& \textbf{10.66} &\textbf{0.78} &\textbf{5.22} & {19.32}& {0.49} & &\textbf{54.20}& \textbf{6.98}&\textbf{0.89} &\textbf{0.18} & \textbf{0.02}& \textbf{0.48}& & \textbf{38.08}& {10.34}& \textbf{0.79}& \textbf{0.70}& \textbf{0.29} & \textbf{0.45}\\
& &\footnotesize{B} & &26.86 &10.71 &0.72 &6.62&\textbf{16.21} &\textbf{0.54}& &29.68 & 8.64& 0.79 & 1.19& 0.52 &0.47& &20.03 & \textbf{7.42}& 0.72& 1.65 & 1.80& 0.36\\
 \cline{5-10} \cline{12-17} \cline{19-24}
&\multirow{2}*{\scriptsize{$\textrm{FC}_{3}$}}&\footnotesize{S} & & \textbf{34.53}&{14.21}& \textbf{0.72} & \textbf{4.15} & \textbf{4.55} & \textbf{0.50} & &$-$&$-$ &$-$&$-$&$-$&$-$ & & $-$ &$-$&$-$ &$-$&$-$&$-$\\
& &\footnotesize{B} & &24.53 & \textbf{10.95} & 0.69 & 20.66& 95.29 & 0.49& &$-$&$-$ &$-$&$-$&$-$&$-$ & & $-$ &$-$&$-$ &$-$&$-$&$-$\\
\cline{1-3} \cline{5-10} \cline{12-17} \cline{19-24}

 \multirow{8}*{\footnotesize{VGG-16}}&
  \multirow{2}*{\scriptsize{$\textrm{Conv}$}}&\footnotesize{S}  & 
  & \textbf{37.00} & 10.87 & \textbf{0.81} & \textbf{0.76} & \textbf{0.40} & \textbf{0.56}& 
  &$-$&$-$ &$-$&$-$&$-$&$-$ & 
  & \textbf{46.55} & 12.73& \textbf{0.79} &  \textbf{0.85}& \textbf{0.80}&\textbf{0.49}  \\
& &\footnotesize{B} & 
&25.94 & \textbf{9.67} & 0.74 &1.81 & 3.04 & 0.52& 
&$-$&$-$ &$-$&$-$&$-$&$-$ & 
&  33.34&\textbf{9.23}  & 0.78 & 1.51 & 1.11 & 0.44 \\
\cline{5-10} \cline{12-17} \cline{19-24}
& \multirow{2}*{\scriptsize{$\textrm{FC}_{1}$}}&\footnotesize{S}  & & \textbf{43.77} & \textbf{8.73} & \textbf{0.84} & \textbf{0.64} & \textbf{0.06} & \textbf{0.66}& & \textbf{56.29}& \textbf{3.13}  &\textbf{0.95} & \textbf{0.02}& \textbf{0.0001}& \textbf{0.47}& & \textbf{42.26} & {11.54} & \textbf{0.80} & \textbf{0.33} & \textbf{0.09} & \textbf{0.52}\\
& &\footnotesize{B} & &22.50 & 11.27 & 0.68 &2.38 & 4.98 & 0.50& &36.06 & 7.71& 0.83& 0.40& 0.13 & 0.44& & 26.87 & \textbf{8.26} & 0.76 & 1.65 & 0.61 & 0.48\\
\cline{5-10} \cline{12-17} \cline{19-24}
&\multirow{2}*{\scriptsize{$\textrm{FC}_{2}$}}&\footnotesize{S}  & &\textbf{36.83} &  \textbf{11.03} & \textbf{0.77} &\textbf{0.80} & \textbf{0.37} & \textbf{0.54}& &{37.79} &  \textbf{4.31}& \textbf{0.90}& \textbf{0.17}& \textbf{0.02}& {0.32}& & \textbf{31.19} & {8.70} & {0.78} & \textbf{0.83} & \textbf{0.45} & {0.35}\\
& &\footnotesize{B} & &23.31 &11.56 &0.67 &5.43 & 22.96 &0.50& &\textbf{38.41} & 9.66 &0.80& 0.79& 0.52&\textbf{0.43}& & 29.37 & \textbf{8.04} & \textbf{0.78} & 2.65 & 1.90 & \textbf{0.46}\\
\cline{5-10} \cline{12-17} \cline{19-24}
&\multirow{2}*{\scriptsize{$\textrm{FC}_{3}$}}&\footnotesize{S}  & & \textbf{32.32} & {10.21} & \textbf{0.77} & \textbf{6.17} & \textbf{32.63} & \textbf{0.47}& &$-$&$-$ &$-$&$-$&$-$&$-$ & & $-$ &$-$&$-$ &$-$&$-$&$-$\\
& &\footnotesize{B} & &23.26 & \textbf{9.97} & 0.71 &17.53 & 216.05 & 0.46& &$-$&$-$ &$-$&$-$&$-$&$-$ & & $-$ &$-$&$-$ &$-$&$-$&$-$\\
\cline{1-3} \cline{5-10} \cline{12-17} \cline{19-24}

\multirow{8}*{\footnotesize{VGG-19}}
&\multirow{2}*{\scriptsize{$\textrm{Conv}$}}&\footnotesize{S}  & & \textbf{40.03}&  \textbf{8.73}& \textbf{0.83} &\textbf{1.59} & \textbf{0.72} & \textbf{0.51}& 
&$-$&$-$ &$-$&$-$&$-$&$-$ &
& \textbf{46.93}  & 14.02& \textbf{0.78} & \textbf{0.41} & \textbf{0.2} &\textbf{0.46} \\
& &\footnotesize{B} & &23.42 &10.58 & 0.69 & 1.81 &3.28 &0.46& 
&$-$&$-$ &$-$&$-$&$-$&$-$ & 
&31.68  &  \textbf{10.17}&0.76  &  1.10& 0.98 &0.39\\
\cline{5-10} \cline{12-17} \cline{19-24}
&\multirow{2}*{\scriptsize{$\textrm{FC}_{1}$}}&\footnotesize{S}  & & \textbf{40.74}&  \textbf{10.42}& \textbf{0.80} &\textbf{0.66} & \textbf{0.15} & \textbf{0.60}& & \textbf{46.50}&  \textbf{2.52}& \textbf{0.95}&\textbf{0.16} & \textbf{0.0002}& \textbf{0.39}& &\textbf{46.38}& {14.05}& {0.77} & \textbf{0.25}& \textbf{0.07}& \textbf{0.45} \\
& &\footnotesize{B} & &22.42 &11.19 & 0.67 & 2.33 &3.67 &0.47& &29.71 &5.83& 0.84& 0.33& 0.12 &0.39& & 28.65&  \textbf{7.93}& \textbf{0.78}& 1.10& 0.80& 0.41\\
\cline{5-10} \cline{12-17} \cline{19-24}
&\multirow{2}*{\scriptsize{$\textrm{FC}_{2}$}}&\footnotesize{S}  & & \textbf{40.20} & \textbf{9.03} & \textbf{0.82} & \textbf{1.16} & \textbf{0.63} & \textbf{0.56}& & \textbf{50.90}& \textbf{5.96} & \textbf{0.91} & \textbf{0.06}& \textbf{0.0006}& \textbf{0.37}& & \textbf{47.03}& {13.66}  & \textbf{0.78}& \textbf{0.10} & \textbf{0.03} & \textbf{0.45}\\
& &\footnotesize{B} & &24.00 &10.40 & 0.70 & 4.64 & 19.07 & 0.47& & 30.31 & 6.15 & 0.84 & 0.45 & 0.18 & 0.37& & 28.46& \textbf{8.20} & 0.78 & 2.14 & 1.92 & 0.41\\
\cline{5-10} \cline{12-17} \cline{19-24}
&\multirow{2}*{\scriptsize{$\textrm{FC}_{3}$}}&\footnotesize{S}  & &\textbf{28.60} &  \textbf{6.37}& \textbf{0.82}& \textbf{4.89}& \textbf{11.57}& \textbf{0.48} & &$-$&$-$ &$-$&$-$&$-$&$-$ & & $-$ &$-$&$-$ &$-$&$-$&$-$\\
& &\footnotesize{B} & &21.29 & 7.77 & 0.74 & 20.61 &143.61 & 0.46& &$-$&$-$ &$-$&$-$&$-$&$-$ & & $-$ &$-$&$-$ &$-$&$-$&$-$\\
\cline{1-3} \cline{5-10} \cline{12-17} \cline{19-24}

\multirow{6}*{\scriptsize{ResNet-50}}&\multirow{2}*{\scriptsize{$\textrm{FC}_{1}$}}&\footnotesize{S} & & \textbf{43.02}&  \textbf{10.15}&\textbf{0.81} & 24.43& 166.76& 0.48& & \textbf{56.00}&  {6.50}&\textbf{0.90} & 3.45& \textbf{4.74}& \textbf{0.45}& & \textbf{42.54}& {10.76}&0.80 & 3.43& 19.60& \textbf{0.40}\\
& &\footnotesize{B} & &42.15 &11.83 &0.79&\textbf{20.78}& \textbf{122.79}&\textbf{0.53}& &43.80 &\textbf{5.75} &0.89&\textbf{2.73}& 6.82&0.36& &39.65 &\textbf{9.81} &\textbf{0.81}&\textbf{1.64}& \textbf{15.20}&0.39\\
\cline{5-10} \cline{12-17} \cline{19-24}
&\multirow{2}*{\scriptsize{$\textrm{FC}_{2}$}}&\footnotesize{S} & &\textbf{48.58}&  \textbf{9.75}&\textbf{0.83} &37.62& \textbf{206.22}& \textbf{0.55}& &\textbf{52.57}&  \textbf{6.54}&\textbf{0.90} &0.25& {1.45}& \textbf{0.40}& &\textbf{41.03}& {12.37}&{0.77} &\textbf{1.85}& \textbf{13.03}& \textbf{0.41}\\
& &\footnotesize{B} & &42.06 &11.88 &0.79&\textbf{29.28}& 248.03&0.52& &43.63 &6.93 &0.87&\textbf{0.02}& \textbf{0.02}&0.35& &38.00 &\textbf{10.00} &\textbf{0.80}&2.68& 30.91&0.38\\
\cline{5-10} \cline{12-17} \cline{19-24}
&\multirow{2}*{\scriptsize{$\textrm{FC}_{3}$}}&\footnotesize{S}  & & {41.38}  &{11.73}  &0.77  &926.61  &\scriptsize{142807.00}  &0.43 & &$-$&$-$ &$-$&$-$&$-$&$-$ & & $-$ &$-$&$-$ &$-$&$-$&$-$\\
& &\footnotesize{B} & &\textbf{42.03}  &\textbf{11.48}  &\textbf{0.79} &\textbf{111.18}  &\textbf{3299.20}  &\textbf{0.53}& &$-$&$-$ &$-$&$-$&$-$&$-$ & & $-$ &$-$&$-$ &$-$&$-$&$-$\\
\cline{1-3} \cline{5-10} \cline{12-17} \cline{19-24}

\multirow{6}*{\scriptsize{ResNet-101}}&\multirow{2}*{\scriptsize{$\textrm{FC}_{1}$}}&\footnotesize{S} & & \textbf{45.93}&  \textbf{11.14}&\textbf{0.81} & \textbf{23.32}& \textbf{236.76}& 0.51& & \textbf{48.59}&  \textbf{5.06}&\textbf{0.91} & \textbf{1.99}& \textbf{2.20}& \textbf{0.39}& & {42.54}&  {9.37}&{0.82} & \textbf{1.39}& \textbf{32.87}& {0.35}\\
& &\footnotesize{B} & &44.18 &12.55 &0.78&40.41& 828.72&\textbf{0.52}& &42.94 &8.16 &0.84&{5.41}& 10.39&0.35& &\textbf{43.33} &\textbf{9.30} &\textbf{0.83}&{15.28}& {48.71}&\textbf{0.39}\\
\cline{5-10} \cline{12-17} \cline{19-24}
&\multirow{2}*{\scriptsize{$\textrm{FC}_{2}$}}&\footnotesize{S} & &\textbf{51.59}&  \textbf{9.02}&\textbf{0.85} &67.60& \textbf{947.85}& \textbf{0.54}& &\textbf{49.27}&  \textbf{6.39}&\textbf{0.89} &\textbf{0.98}& \textbf{0.65}& \textbf{0.37}& &\textbf{41.71}&  {9.16}&{0.82} &{3.30}& {100.97}& {0.38}\\
&&\footnotesize{B} & &43.22 &12.32 &0.78&\textbf{43.40}& \small{1155.22}&0.50& &41.79 &7.30 &0.85&{6.58}& {17.16}&0.34& &41.35 &\textbf{8.32} &\textbf{0.84}&\textbf{2.26}& \textbf{48.61}&\textbf{0.39}\\
 \cline{5-10} \cline{12-17} \cline{19-24}
&\multirow{2}*{\scriptsize{$\textrm{FC}_{3}$}}&\footnotesize{S}  & &\textbf{47.71}  &\textbf{10.24}  &\textbf{0.82}  &\textbf{73.33}  &\textbf{2797.15}  &\textbf{0.53}  & &$-$&$-$ &$-$&$-$&$-$&$-$ & & $-$ &$-$&$-$ &$-$&$-$&$-$\\
& &\footnotesize{B} & &42.40  &10.53  &{0.80} &{162.68}  &\footnotesize{16481.93}  &0.49& &$-$&$-$ &$-$&$-$&$-$&$-$ & & $-$ &$-$&$-$ &$-$&$-$&$-$\\
\cline{1-3} \cline{5-10} \cline{12-17} \cline{19-24}

\multirow{6}*{\scriptsize{ResNet-152}}&\multirow{2}*{\scriptsize{$\textrm{FC}_{1}$}}&\footnotesize{S} & & {44.81}& {12.09}&{0.79} & \textbf{26.35}& \textbf{289.59}& 0.48& & \textbf{44.90}&  {5.63}&\textbf{0.89} & {6.25}& \textbf{5.86}& \textbf{0.36}& & \textbf{41.09}&  \textbf{10.09}&\textbf{0.81} & \textbf{0.33}& \textbf{3.59}& \textbf{0.39}\\
& &\footnotesize{B} & &\textbf{45.62} &\textbf{10.68} &\textbf{0.81}&36.92& 767.58&\textbf{0.54}& &39.93 &\textbf{5.40} &0.89&\textbf{6.08}& 6.74&0.33& &40.15 &10.82 &{0.79}&{0.59}& {11.39}&{0.37}\\
\cline{5-10} \cline{12-17} \cline{19-24}
&\multirow{2}*{\scriptsize{$\textrm{FC}_{2}$}}&\footnotesize{S} & &{43.79}&  \textbf{10.04}&\textbf{0.81} &\textbf{7.13}& \textbf{42.77}& {0.52}& &\textbf{40.98}&  {6.90}&{0.86} &\textbf{4.64}& \textbf{5.71}& {0.32}& &\textbf{41.36}& \textbf{12.04}&\textbf{0.78} &\textbf{14.29}& \textbf{17.33}& \textbf{0.38}\\
&&\footnotesize{B} & &\textbf{45.08} &10.85 &0.81&{44.59}& \small{1200.97}&\textbf{0.52}& &{40.29} &\textbf{5.56} &\textbf{0.89}&{7.86}& {12.24}&\textbf{0.33}& &38.57 &12.07 &{0.77}&{18.03}& {67.52}&{0.36}\\
\cline{5-10} \cline{12-17} \cline{19-24}
&\multirow{2}*{\scriptsize{$\textrm{FC}_{3}$}}&\footnotesize{S}  & &{44.21}  &{11.89}  &{0.79}  &\textbf{47.28}  &\textbf{1463.55}  &{0.50}& &$-$&$-$ &$-$&$-$&$-$&$-$ & & $-$ &$-$&$-$ &$-$&$-$&$-$\\
& &\footnotesize{B} & &\textbf{44.89}  &\textbf{10.77}  &\textbf{0.81} &{167.41}  &\footnotesize{16331.28}  &\textbf{0.52}& &$-$&$-$ &$-$&$-$&$-$&$-$ & & $-$ &$-$&$-$ &$-$&$-$&$-$\\

\bottomrule
\end{tabular}}
\end{center}
\vspace{-10pt}
\end{table*}

\begin{table}[t]
\begin{center}
\caption{Comparisons between the student network (S) and the baseline network (B) for the question-answering and the sentiment classification tasks, respectively. Metrics were evaluated at the last hidden layer of the BERT model, which verified Hypotheses 1-3.}
\label{tab:NLP}
\resizebox{0.98\linewidth}{!}{\
\begin{tabular}{p{0.2\linewidth}<{\centering}| p{0.2\linewidth}<{\centering} p{0.05\linewidth}<{\centering}|p{0.1\linewidth}<{\centering} p{0.1\linewidth}<{\centering}p{0.1\linewidth}<{\centering}| p{0.1\linewidth}<{\centering} p{0.1\linewidth}<{\centering}|p{0.05\linewidth}<{\centering}}
\toprule
Dataset &Network&&$N_{\textrm{point}}^{\textrm{fg}}\uparrow$& $N_{\textrm{point}}^{\textrm{bg}}\downarrow$ & $\lambda\uparrow$ &$D_{\textrm{mean}}\downarrow$& $D_{\textrm{var}}\downarrow$&$\rho\uparrow$\\
\midrule

 \multirow{2}*{SQuAD}&\multirow{2}*{BERT }
  &{S} &  {\textbf{8.46}}  &  {\textbf{10.24}}&  {\textbf{0.48}}& \textbf{13.81}& \textbf{104.94}& \textbf{0.56}\\
& &{B} & 3.58  & {10.55} &0.29 &23.32 & 206.23&0.25 \\
\midrule

 \multirow{2}*{SST-2}&\multirow{2}*{BERT}
  &{S} &{ \textbf{1.31}}  & {1.13}&  {\textbf{0.53}}& \textbf{1.08}& \textbf{0.68}& \textbf{0.80}\\
& &{B} & 0.68  & { \textbf{0.79}} &0.39 & 1.29&1.00 & 0.51 \\
\bottomrule
\end{tabular}}
\end{center}
\vspace{-15pt}
\end{table}

\begin{table}[t]
\begin{center}
\caption{Comparisons between the student network (S) and the baseline network (B) for the classification of 3D point clouds. In general, Hypotheses 1-3 were verified.}
\label{tab:pointcloud}
\resizebox{0.98\linewidth}{!}{\
\begin{tabular}{p{0.2\linewidth}<{\centering}| p{0.1\linewidth}<{\centering} p{0.05\linewidth}<{\centering}|p{0.12\linewidth}<{\centering} p{0.12\linewidth}<{\centering}p{0.08\linewidth}<{\centering}| p{0.12\linewidth}<{\centering} p{0.12\linewidth}<{\centering}|p{0.05\linewidth}<{\centering}}
\toprule
Network &Layer&&$N_{\textrm{point}}^{\textrm{fg}}\uparrow$& $N_{\textrm{point}}^{\textrm{bg}}\downarrow$ & $\lambda\uparrow$ &$D_{\textrm{mean}}\downarrow$& $D_{\textrm{var}}\downarrow$&$\rho\uparrow$\\
\midrule

 \multirow{4}*{PointNet}&\multirow{2}*{$\textrm{FC}_{1}$}
  &{S} & {\textbf{157.36}} &63.62&   {\textbf{0.72}}& \textbf{17.29}& \textbf{86.39}& \textbf{0.29}\\
& &{B} &122.59   &  {\textbf{52.67}} &0.71 & 54.88 & 595.03 &0.28 \\
\cline{2-9}
&\multirow{2}*{$\textrm{FC}_{2}$}
  &{S} &{ \textbf{131.91}}  & 56.51& 0.70& \textbf{21.58}& \textbf{134.09}& \textbf{0.28}\\
& &{B} &  120.50 & {\textbf{50.36}}& {\textbf{0.72}} & 39.15 & 507.11 & 0.27 \\
\midrule

\multirow{4}*{DGCNN}&\multirow{2}*{$\textrm{FC}_{1}$}
  &{S} &84.44  & {\textbf{30.78}}&  {\textbf{0.79}}&  \textbf{0.83}& \textbf{3.08}& \textbf{0.48}\\
& &{B} &  { \textbf{102.53}}  & 36.04& 0.79& 6.16& 5.51& 0.41 \\
\cline{2-9}
&\multirow{2}*{$\textrm{FC}_{2}$}
  &{S} & 54.25  &{ \textbf{13.71}}& {\textbf{0.87}}& \textbf{1.25}& \textbf{1.05}& \textbf{0.55}\\
& &{B} & {\textbf{119.08}}  &31.10 & 0.82& 4.66& 5.46& 0.33 \\
\midrule

\multirow{4}*{PointConv}&\multirow{2}*{$\textrm{FC}_{1}$}
  &{S} &  {\textbf{73.09}}  & 8.71&  {\textbf{0.91}}& {\textbf{0.08}}& \textbf{0.01}& \textbf{0.10}\\
& &{B} &  70.14 & {\textbf{7.98}}& 0.89& 0.95& 0.10&0.07 \\
\cline{2-9}
&\multirow{2}*{$\textrm{FC}_{2}$}
  &{S} & 66.08  &{ \textbf{9.45}}&0.89 & \textbf{0.06}& \textbf{0.003}& \textbf{0.13}\\
& &{B} & { \textbf{83.37 }}&10.68 &{ \textbf{0.89}}&0.83 & 0.03&0.05 \\
\midrule

\multirow{4}*{PointNet++}&\multirow{2}*{$\textrm{FC}_{1}$}
  &{S} & { \textbf{51.15} } & 8.17&  {\textbf{0.83}}& \textbf{1.98}& \textbf{2.24}& 0.33\\
& &{B} &  16.04 & { \textbf{6.23}}& 0.64& 3.45& 4.15&\textbf{0.62} \\
\cline{2-9}
&\multirow{2}*{$\textrm{FC}_{2}$}
  &{S} & { \textbf{50.88}}  & {\textbf{13.24}}& {\textbf{0.83}}&2.12& \textbf{2.20}& 0.25\\
& &{B} &  42.53 &13.75&0.81& \textbf{1.81} & 2.33&\textbf{0.31} \\
\bottomrule
\end{tabular}}
\end{center}
\vspace{-10pt}
\end{table}

\begin{table}[t]
\begin{center}
\caption{Comparisons between the student network (S) and the baseline network (B) for image classification on the MSCOCO dataset. Metrics were evaluated at $\textrm{FC}_{3}$ layer.}
\label{tab:coco}
\resizebox{0.98\linewidth}{!}{\
\begin{tabular}{p{0.15\linewidth}<{\centering}| p{0.1\linewidth}<{\centering} p{0.05\linewidth}<{\centering}|p{0.1\linewidth}<{\centering} p{0.1\linewidth}<{\centering}| 
p{0.15\linewidth}<{\centering}| p{0.1\linewidth}<{\centering} p{0.05\linewidth}<{\centering}|p{0.1\linewidth}<{\centering} p{0.1\linewidth}<{\centering}
}
\toprule
Network & Layer & &$N_{\textrm{point}}^{\textrm{fg}}$ & $N_{\textrm{point}}^{\textrm{bg}}$ & 
Network& Layer & &$N_{\textrm{point}}^{\textrm{fg}}$& $N_{\textrm{point}}^{\textrm{bg}}$ \\
\midrule

\multirow{2}*{AlexNet}&\multirow{2}*{$\textrm{FC}_{3}$}
  &{S} & 41.04  &  1.52 &
  \multirow{2}*{VGG-11}&\multirow{2}*{$\textrm{FC}_{3}$}
  &{S} &  30.96  & 7.16    \\
  
  & &{B} & 41.44   &1.88 & 
& &{B} &36.20  & 7.20  \\
\midrule

 \multirow{2}*{VGG-16}&\multirow{2}*{$\textrm{FC}_{3}$}
  &{S} & 31.35  &  6.93&
  \multirow{2}*{VGG-19}&\multirow{2}*{$\textrm{FC}_{3}$}
  &{S} & 31.70  & 7.11 \\
  
& &{B} &  35.09  & 7.02   &
& &{B} & 35.23  & 7.11  \\
\bottomrule
\end{tabular}}
\end{center}
\vspace{-15pt}
\end{table}


\subsection{Verification of Hypothesis 1}
\label{4.2}
According to Hypothesis 1, a well-trained teacher network was supposed to encode more foreground knowledge points than the student network and the baseline network. 
Consequently, a distilled student network was supposed to model more foreground knowledge points than the baseline network.
This was the case because the teacher network was usually trained using a large amount of training data, and achieved superior performance to the baseline network.
In this way, this well-trained teacher network was supposed to encode more foreground knowledge points than the baseline network. 
Knowledge distillation forced the student network to mimic the teacher network.
Hence, the student network was supposed to model more foreground knowledge points than baseline network.

$\bullet\quad$\textbf{Quantification of knowledge points in the teacher network, the student network, and the baseline network.}
Here, we compared knowledge points encoded in the teacher network, the student network, and the baseline network, and verified the above hypothesis.
We trained a VGG-16 model from scratch as the teacher network using the ILSVRC-2013 DET dataset or the CUB200-2011 dataset for image classification. 
Data augmentation~\cite{jacobsen2018revnet} was used to boost the performance of the teacher network.
Table~\ref{teacher supplementary} compares the number of foreground knowledge points $N_{\textrm{point}}^{\textrm{fg}}$ learned by the teacher network, the baseline network, and the student network.
For fair comparisons, the student network had the same architecture as the teacher network and the baseline network according to Section \ref{Implementation Details}.

Table~\ref{teacher supplementary} shows that the teacher network usually encoded more foreground knowledge points $N_{\textrm{point}}^{\textrm{fg}}$ and a higher ratio $\lambda$ than the student network.
Meanwhile, the student network often obtained larger values of $N_{\textrm{point}}^{\textrm{fg}}$ and $\lambda$ than the baseline network.
In this way, the above hypothesis was verified.
There was an exception when $N_{\textrm{point}}^{\textrm{fg}}$ was measured at the $\textrm{FC}_{2}$ layer of the VGG-16 model on the ILSVRC-2013 DET dataset.
The teacher network encoded fewer foreground knowledge points than the student network, because this teacher network had a larger average background entropy value $\overline{H}$ than the student network.

$\bullet\quad$\textbf{Comparing metrics of $N_{\textrm{point}}^{\textrm{fg}}$, $N_{\textrm{point}}^{\textrm{bg}}$ and $\lambda$ to verify Hypothesis 1.}
In contrast to the previous subsection learning the teacher network from scratch, here, we compared $N_{\textrm{point}}^{\textrm{fg}}$, $N_{\textrm{point}}^{\textrm{bg}}$ and $\lambda$ between the student network and the baseline network when using a teacher network trained in a more sophisticated manner.
In other words, this teacher network was pre-trained on the ImageNet dataset, and then fine-tuned using the ILSVRC-2013 DET dataset, the CUB200-2011 dataset, or the Pascal VOC 2012 dataset. 
This is a more common case in real applications.

Based on the information discarding $\{H_{i}\}$, we visualized image regions in the foreground that corresponded to foreground knowledge points, as well as image regions in the background that corresponded to background knowledge points in Figure~\ref{fig4}, where these knowledge points were measured at the $\textrm{FC}_{1}$ layer of the VGG-11 model.
In this scenario, Hypothesis 1 was also successfully verified.
Moreover, for tasks of image classification, natural language processing, and 3D point cloud classification, 
Table~\ref{tab:image class}, Table~\ref{tab:NLP}, and Table~\ref{tab:pointcloud} show that Hypothesis 1 was vaildated.
Very few student networks in Table~\ref{tab:image class}, Table~\ref{tab:NLP}, and Table~\ref{tab:pointcloud} encoded more background knowledge points $N_{\textrm{point}}^{\textrm{bg}}$ than the baseline network. 
This occurred because teacher networks in this subsection, Section~\ref{Verification of Hypothesis 2}, and Section~\ref{Verification of Hypothesis 3} were either pre-trained or trained using more samples, which encoded far more knowledge points than necessary.
In this way, student networks distilled from these well-trained teacher network learned more unnecessary knowledge for inference, thereby leading to a larger $N_{\textrm{point}}^{\textrm{bg}}$ value than the baseline network.

Note that as discussed in Section~\ref{sec:Algorithm}, knowledge distillation usually involved two different utilities.
The utility of distilling from the low-dimensional network output was mainly to select confident training samples for learning, with very little information on how the teacher network encoded detailed features.
In contrast, the utility of distilling from high-dimensional intermediate-layer features was mainly to force the student network to mimic knowledge points of the teacher network.
Hence, in this study, we mainly explained knowledge distillation from high-dimensional intermediate-layer features.

To this end, in Table~\ref{tab:image class}, we mainly focused on distillation from intermediate-layer features.
Nevertheless, we conducted another experiment to verify that the utility of distilling from relatively low-dimensional network output was to select confident samples and ignore unconfident samples.
Specifically, we distilled the student network from the $10-$dimensional final output of the teacher network for the classification of the top-10 largest categories in the MSCOCO dataset~\cite{lin2014microsoft}.
Results in Table~\ref{tab:image class} and Table~\ref{tab:coco} show the difference between the two utilities of knowledge distillation, to some extent.
For distilling from the 200-dimensional output for the classification on the CUB200-2011 dataset in Table~\ref{tab:image class}, the Hypothesis 1 was verified, because the network output with $200$ dimensions encoded relatively rich information to force the student network to mimic knowledge points of the teacher network, which reflected the first utility of knowledge distillation.
In contrast, when the network output had very few dimensions, such as 10-category classification on the MSCOCO dataset in Table~\ref{tab:coco},  the student network tended to encode both fewer foreground knowledge points and fewer background knowledge points than learning from scratch.
This phenomenon verified the second utility of sample selection.

Additionally, unlike $\textrm{FC}$ layers, convolutional layers were usually not sufficiently fine-tuned due to their depth, when the teacher networks were first pre-trained on the ImageNet dataset for 1000-category classification, and were then fine-tuned using the CUB200-2011 dataset or the Pascal VOC 2012  dataset.
In this scenario, convolutional layers of the teacher network might encode some noisy features, such as features pre-trained for massive unrelated ImageNet categories.
This might violate the first requirement for the knowledge distillation~\textit{i.e.,} the teacher network was supposed to be well optimized for the same task as the student network.
Hence, in order to analyze the convolutional layer in a stricter manner, we conducted another experiment.
Specifically, we fine-tuned the teacher network on the Tiny ImageNet dataset~\cite{le2015tiny}, which was much larger and already contained sufficient
training images to ensure that features in high convolutional layers were fine-tuned in a sophisticated manner. 
Table~\ref{tab:tiny} shows that most hypotheses were verified, except that the student network might encode more background knowledge points than the baseline network.
This was the case because the teacher network encoded significantly more foreground knowledge points in convolutional layers than the baseline network. 
As a cost, the teacher network also encoded slightly more background knowledge points than the baseline network. 
Therefore, the student network encoded much more foreground knowledge points than the baseline network ($3.9$ more foreground knowledge points on average), but simply increased background knowledge points by $1.1$.
Nevertheless, the student network still exhibited a larger ratio of foreground knowledge points than the baseline network.
We considered that this phenomenon was still reasonable, as explained above.

\textbf{Correlation between the quality of knowledge points and classification performance.}
We conducted experiments to check whether the DNN encoding a larger ratio of foreground knowledge points $\lambda$ usually achieved better classification performance, as discussed in Section~\ref{hypothesis1}.
To this end, we used the aforementioned student network to describe how the ratio $\lambda$ changes in different epochs of the training process, which was distilled from $\textrm{FC}_1$ layer of the teacher network on the CUB200-2011 dataset for object classification.
Figure~\ref{fig:acc_know} shows that DNNs with a larger ratio of task-related knowledge points often exhibited better performance.


\begin{figure*}[t]
    \centering
    \includegraphics[width=0.9\linewidth]{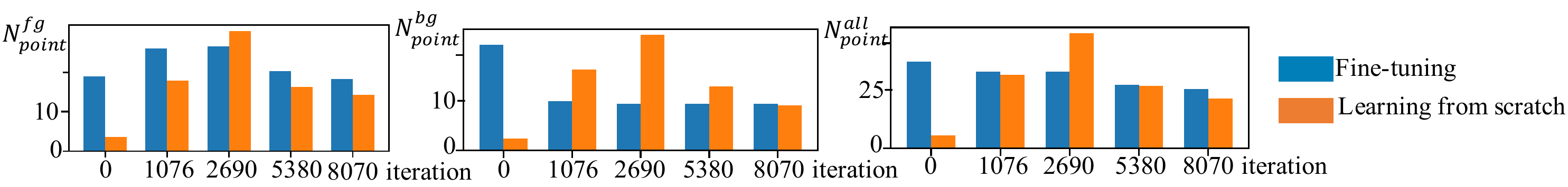}
    \vspace{-5pt}
    \caption{Comparison of {\small{$N_{\textrm{point}}^{\textrm{fg}}$}}, {\small{$N_{\textrm{point}}^{\textrm{bg}}$}} and all knowledge points {\small{$N_{\textrm{point}}^{\textrm{all}}= N_{\textrm{point}}^{\textrm{fg}}+ N_{\textrm{point}}^{\textrm{bg}}$}} between fine-tuning a pre-trained VGG-16 network and learning a VGG-16 network from scratch on the Tiny-ImageNet dataset. Metrics were measured at the highest convolutional layer. The fine-tuning process made the DNN encode new foreground knowledge points more quickly and 
   enabled it to be optimized with fewer detours than learning from scratch.}
    \label{transfer}
    \vspace{-10pt}
\end{figure*}


\begin{table}[t]
\begin{center}
\caption{Comparisons between the student network (S) and the baseline network (B) for image classification on the Tiny-ImageNet dataset. Metrics were evaluated at $\textrm{Conv}$ layer, where Hypotheses 1-3 were verified, to some extent.}
\label{tab:tiny}
\resizebox{0.98\linewidth}{!}{\
\begin{tabular}{p{0.15\linewidth}<{\centering}| p{0.35\linewidth}<{\centering} |p{0.1\linewidth}<{\centering} p{0.1\linewidth}<{\centering}p{0.05\linewidth}<{\centering}| p{0.1\linewidth}<{\centering} p{0.1\linewidth}<{\centering}|p{0.05\linewidth}<{\centering}}
\toprule
\small{Target network}&\small{\makecell[c]{Learning\\ methods}}&$N_{\textrm{point}}^{\textrm{fg}}\uparrow$& $N_{\textrm{point}}^{\textrm{bg}}\downarrow$ & $\lambda\uparrow$ &$D_{\textrm{mean}}\downarrow$& $D_{\textrm{var}}\downarrow$&$\rho\uparrow$\\
\midrule

 \multirow{2}*{AlexNet}
&\scriptsize{Distilling from AlexNet}  & \textbf{14.45}  &5.88 &  \textbf{0.67} & \textbf{1.51}& \textbf{3.09}&0.32 \\
& \footnotesize{Learning from scratch}  & 10.14  & \textbf{5.14} & 0.66 & 24.98& 586.45 & \textbf{0.43} \\
\midrule

 \multirow{4}*{VGG-11}
  &\footnotesize{Distilling from VGG-11}&  {\textbf{19.78}}  & 9.60 & \textbf{0.65}& \textbf{2.58}& \textbf{13.41}& \textbf{0.37}\\
& \footnotesize{Learning from scratch} & 14.33 & \textbf{8.18} & 0.61& 6.58& 131.92 & 0.27 \\
\cline{2-8}

&\footnotesize{Distilling from VGG-16} &  {\textbf{17.42}}  & 9.96 & \textbf{0.63}& \textbf{4.47}& \textbf{22.23}& \textbf{0.32}\\
&  \footnotesize{Learning from scratch}  & 14.33 & \textbf{8.18} & 0.61& 6.58& 131.92 & 0.27 \\
\midrule


 \multirow{4}*{VGG-16}
  &\footnotesize{Distilling from VGG-16} &  {\textbf{17.76}}  & 9.33 & 0.63 & \textbf{0.85}& \textbf{0.93}& \textbf{0.32}\\
&  \footnotesize{Learning from scratch}  &  15.29& \textbf{7.79} &\textbf{0.65}  & 3.46 & 40.31 & 0.27 \\
\cline{2-8}

&\footnotesize{Distilling from VGG-19}&  {\textbf{19.88}}  & 8.10 & \textbf{0.71}& 5.93 & \textbf{22.08}& \textbf{0.44}\\
&  \footnotesize{Learning from scratch} & 15.29 &  \textbf{7.79} & 0.65&  \textbf{3.46}& 40.31 & 0.27 \\
\midrule

 \multirow{2}*{VGG-19}
   &\footnotesize{Distilling from VGG-19}& 15.91  &\textbf{8.67}& 0.61 & \textbf{0.79}& \textbf{0.88}& \textbf{0.26}\\
&  \footnotesize{Learning from scratch} & \textbf{16.78}  & 9.16 & \textbf{0.64} & 4.39& 87.42 &  0.26 \\

 \bottomrule
\end{tabular}}
\end{center}
\vspace{-15pt}
\end{table}

%
%

\subsection{Verification of Hypothesis 2}
\label{Verification of Hypothesis 2}
For Hypothesis 2, we assumed that knowledge distillation made the student network learn different knowledge points simultaneously. To this end, we used metrics $D_{\textrm{mean}}$ and $D_{\textrm{std}}$ to verify this hypothesis.

For image classification,
Table \ref{tab:image class} shows that values of $D_{\textrm{mean}}$ and $D_{\textrm{std}}$ were usually smaller than those of the baseline network, which verified Hypothesis 2.
Of note, a very small number of failure cases occurred.
For example, the $D_{\textrm{mean}}$ and $D_{\textrm{std}}$ were measured at the $\textrm{FC}_{1}$ layer of  AlexNet or at the $\textrm{Conv}$ layer of VGG-11. 
It is because both the AlexNet model and the VGG-11 model had relatively shallow network architectures.
When learning from raw data, DNNs with shallow architectures would more easily learn more knowledge points with less overfitting.

For NLP tasks (question-answering and sentiment classification) and the classification of 3D point clouds, 
Table~\ref{tab:NLP} and Table~\ref{tab:pointcloud} show that the student network had smaller values of $D_{\textrm{mean}}$ and $D_{\textrm{std}}$ than the baseline network, respectively, which successfully verified Hypothesis 2.

\subsection{Verification of Hypothesis 3}
\label{Verification of Hypothesis 3}
Hypothesis 3 assumed that knowledge distillation made the student network optimized with fewer detours than the baseline network.
The metric $\rho$ measuring the stability of optimization directions was used to verify this hypothesis.

For image classification, natural language processing, and 3D point cloud classification,
Table \ref{tab:image class}, Table~\ref{tab:NLP}, and Table~\ref{tab:pointcloud} show that the student network usually exhibited a larger $\rho$ value than the baseline network, which verified Hypothesis 3.
However, there were few failure cases.
When measuring $\rho$ by AlexNet and VGG-11, shallow architectures of these two networks lead to failure cases (please see discussions in Section~\ref{Verification of Hypothesis 2}).
There was also another failure case when we used the PointNet++ model to measure $\rho$, because this model was sufficiently discriminative.
In this way, this PointNet++ model could make inferences based on a few knowledge points.


\subsection{Analysis of fine-tuning process}
\label{sec:fine-tune}
We further checked whether knowledge points could be used to analyze other classic transfer-learning methods.
Here, we focused on fine-tuning a well-trained DNN on a new dataset.
Specifically, we conducted experiments to explore how the fine-tuning process guided the DNN to encode new knowledge points, in which we fine-tuned a pre-trained VGG-16 network  on the Tiny ImageNet dataset.
Besides, we also trained another VGG-16 network from scratch on the same dataset  for comparison.
Figure~\ref{transfer} shows that the fine-tuning process made the DNN learn new knowledge points in the foreground more quickly and be optimized with fewer detours than learning from scratch,~\textit{i.e.,} the fine-tuned DNN discarded fewer temporarily learned knowledge points.

\section{Conclusion and discussions}
In this study, we provide a new perspective to explain knowledge distillation from intermediate-layer features,
\textit{i.e.} quantifying knowledge points encoded in the intermediate layers of a DNN.
We propose three hypotheses regarding the mechanism of knowledge distillation, and design three types of metrics to verify these hypotheses for different classification tasks.
Compared to learning from scratch, knowledge distillation ensures that the DNN encodes more knowledge points, learns different knowledge points simultaneously, and optimizes with fewer detours.
Note that the learning procedure of DNNs cannot be precisely divided into a learning phase and a discarding phase. In each epoch, the DNN may simultaneously learn new knowledge points and discard old knowledge points irrelevant to the task. Thus, the target epoch $\hat{m}$ in Figure~\ref{fig2} is simply a rough estimation of the division of two learning phases.

\ifCLASSOPTIONcompsoc
  \section*{Acknowledgments}
\else
  \section*{Acknowledgment}
\fi
This work is partially supported by National Key R\&D Program of China (2021ZD0111602), the National Nature Science Foundation of China (No. 61906120, U19B2043), Shanghai Natural Science Foundation (21JC1403800,21ZR1434600), Shanghai Municipal Science and Technology Major Project (2021SHZDZX0102).
This work is also partially supported by Huawei Technologies Inc.
%

\ifCLASSOPTIONcaptionsoff
  \newpage
\fi

\bibliographystyle{IEEE}
\bibliography{PAMI_distill}

\begin{thebibliography}{10}\itemsep=-1pt

\bibitem{NIPS2017_b22b257a}
P.~L. Bartlett, D.~J. Foster, and M.~J. Telgarsky.
\newblock Spectrally-normalized margin bounds for neural networks.
\newblock In I.~Guyon, U.~V. Luxburg, S.~Bengio, H.~Wallach, R.~Fergus,
  S.~Vishwanathan, and R.~Garnett, editors, {\em Advances in Neural Information
  Processing Systems}, volume~30. Curran Associates, Inc., 2017.

\bibitem{bau2017network}
D.~Bau, B.~Zhou, A.~Khosla, A.~Oliva, and A.~Torralba.
\newblock Network dissection: Quantifying interpretability of deep visual
  representations.
\newblock In {\em Proceedings of the IEEE Conference on Computer Vision and
  Pattern Recognition}, pages 6541--6549, 2017.

\bibitem{behrmann2019invertible}
J.~Behrmann, W.~Grathwohl, R.~T. Chen, D.~Duvenaud, and J.-H. Jacobsen.
\newblock Invertible residual networks.
\newblock In {\em International Conference on Machine Learning}, pages
  573--582. PMLR, 2019.

\bibitem{Brendel19iclr}
W.~Brendel and M.~Bethge.
\newblock Approximating cnns with bag-of-local-features models works
  surprisingly well on imagenet.
\newblock In {\em 7th International Conference on Learning Representations,
  {ICLR} 2019, New Orleans, LA, USA, May 6-9, 2019}. OpenReview.net, 2019.

\bibitem{Chatterji2020The}
N.~Chatterji, B.~Neyshabur, and H.~Sedghi.
\newblock The intriguing role of module criticality in the generalization of
  deep networks.
\newblock In {\em International Conference on Learning Representations}, 2020.

\bibitem{chattopadhay2018grad}
A.~Chattopadhay, A.~Sarkar, P.~Howlader, and V.~N. Balasubramanian.
\newblock Grad-cam++: Generalized gradient-based visual explanations for deep
  convolutional networks.
\newblock In {\em 2018 IEEE Winter Conference on Applications of Computer
  Vision (WACV)}, pages 839--847. IEEE, 2018.

\bibitem{chen2019looks}
C.~Chen, O.~Li, D.~Tao, A.~Barnett, C.~Rudin, and J.~K. Su.
\newblock This looks like that: deep learning for interpretable image
  recognition.
\newblock In {\em Advances in Neural Information Processing Systems}, pages
  8928--8939, 2019.

\bibitem{chen2019residual}
R.~T. Chen, J.~Behrmann, D.~K. Duvenaud, and J.-H. Jacobsen.
\newblock Residual flows for invertible generative modeling.
\newblock {\em Advances in Neural Information Processing Systems}, 32, 2019.

\bibitem{chen2016infogan}
X.~Chen, Y.~Duan, R.~Houthooft, J.~Schulman, I.~Sutskever, and P.~Abbeel.
\newblock Infogan: Interpretable representation learning by information
  maximizing generative adversarial nets.
\newblock In {\em Advances in neural information processing systems}, pages
  2172--2180, 2016.

\bibitem{cheng2020explaining}
X.~Cheng, Z.~Rao, Y.~Chen, and Q.~Zhang.
\newblock Explaining knowledge distillation by quantifying the knowledge.
\newblock In {\em Proceedings of the IEEE/CVF Conference on Computer Vision and
  Pattern Recognition}, pages 12925--12935, 2020.

\bibitem{devlin2018bert}
J.~Devlin, M.-W. Chang, K.~Lee, and K.~Toutanova.
\newblock Bert: Pre-training of deep bidirectional transformers for language
  understanding.
\newblock {\em arXiv preprint arXiv:1810.04805}, 2018.

\bibitem{dinh2017sharp}
L.~Dinh, R.~Pascanu, S.~Bengio, and Y.~Bengio.
\newblock Sharp minima can generalize for deep nets.
\newblock In {\em International Conference on Machine Learning}, pages
  1019--1028. PMLR, 2017.

\bibitem{dosovitskiy2016inverting}
A.~Dosovitskiy and T.~Brox.
\newblock Inverting visual representations with convolutional networks.
\newblock In {\em Proceedings of the IEEE Conference on Computer Vision and
  Pattern Recognition}, pages 4829--4837, 2016.

\bibitem{Everingham10}
M.~Everingham, L.~Van~Gool, C.~K.~I. Williams, J.~Winn, and A.~Zisserman.
\newblock The pascal visual object classes (voc) challenge.
\newblock {\em International Journal of Computer Vision}, 88(2):303--338, June
  2010.

\bibitem{flennerhag2018transferring}
S.~Flennerhag, P.~G. Moreno, N.~D. Lawrence, and A.~Damianou.
\newblock Transferring knowledge across learning processes.
\newblock {\em arXiv preprint arXiv:1812.01054}, 2018.

\bibitem{fong2018net2vec}
R.~Fong and A.~Vedaldi.
\newblock Net2vec: Quantifying and explaining how concepts are encoded by
  filters in deep neural networks.
\newblock In {\em Proceedings of the IEEE Conference on Computer Vision and
  Pattern Recognition}, pages 8730--8738, 2018.

\bibitem{fong2017interpretable}
R.~C. Fong and A.~Vedaldi.
\newblock Interpretable explanations of black boxes by meaningful perturbation.
\newblock In {\em Proceedings of the IEEE International Conference on Computer
  Vision}, pages 3429--3437, 2017.

\bibitem{fort2019stiffness}
S.~Fort, P.~K. Nowak, and S.~Narayanan.
\newblock Stiffness: A new perspective on generalization in neural networks.
\newblock {\em arXiv preprint arXiv:1901.09491}, 2019.

\bibitem{furlanello2018born}
T.~Furlanello, Z.~C. Lipton, M.~Tschannen, L.~Itti, and A.~Anandkumar.
\newblock Born again neural networks.
\newblock {\em arXiv preprint arXiv:1805.04770}, 2018.

\bibitem{garipov2018loss}
T.~Garipov, P.~Izmailov, D.~Podoprikhin, D.~P. Vetrov, and A.~G. Wilson.
\newblock Loss surfaces, mode connectivity, and fast ensembling of dnns.
\newblock In {\em Advances in Neural Information Processing Systems}, pages
  8789--8798, 2018.

\bibitem{goldfeld2019estimating}
Z.~Goldfeld, E.~Van Den~Berg, K.~Greenewald, I.~Melnyk, N.~Nguyen,
  B.~Kingsbury, and Y.~Polyanskiy.
\newblock Estimating information flow in deep neural networks.
\newblock In {\em International Conference on Machine Learning}, pages
  2299--2308, 2019.

\bibitem{gouk2021regularisation}
H.~Gouk, E.~Frank, B.~Pfahringer, and M.~J. Cree.
\newblock Regularisation of neural networks by enforcing lipschitz continuity.
\newblock {\em Machine Learning}, 110(2):393--416, 2021.

\bibitem{guan2019towards}
C.~Guan, X.~Wang, Q.~Zhang, R.~Chen, D.~He, and X.~Xie.
\newblock Towards a deep and unified understanding of deep neural models in
  nlp.
\newblock In {\em International Conference on Machine Learning}, pages
  2454--2463, 2019.

\bibitem{higgins2017beta}
I.~Higgins, L.~Matthey, A.~Pal, C.~Burgess, X.~Glorot, M.~Botvinick,
  S.~Mohamed, and A.~Lerchner.
\newblock beta-vae: Learning basic visual concepts with a constrained
  variational framework.
\newblock {\em ICLR}, 2(5):6, 2017.

\bibitem{hinton2015distilling}
G.~Hinton, O.~Vinyals, and J.~Dean.
\newblock Distilling the knowledge in a neural network.
\newblock {\em arXiv preprint arXiv:1503.02531}, 2015.

\bibitem{hinton2018matrix}
G.~E. Hinton, S.~Sabour, and N.~Frosst.
\newblock Matrix capsules with em routing.
\newblock In {\em International conference on learning representations}, 2018.

\bibitem{hooker2019benchmark}
S.~Hooker, D.~Erhan, P.-J. Kindermans, and B.~Kim.
\newblock A benchmark for interpretability methods in deep neural networks.
\newblock In {\em Advances in Neural Information Processing Systems}, pages
  9737--9748, 2019.

\bibitem{jacobsen2018revnet}
J.-H. Jacobsen, A.~Smeulders, and E.~Oyallon.
\newblock i-revnet: Deep invertible networks.
\newblock {\em arXiv preprint arXiv:1802.07088}, 2018.

\bibitem{he2016deep}
S.~R. Kaiming~He, Xiangyu~Zhang and J.~Sun.
\newblock Deep residual learning for image recognition.
\newblock {\em In {CVPR}}, 2016.

\bibitem{Kapishnikov_2019_ICCV}
A.~Kapishnikov, T.~Bolukbasi, F.~Viegas, and M.~Terry.
\newblock Xrai: Better attributions through regions.
\newblock In {\em Proceedings of the IEEE/CVF International Conference on
  Computer Vision}, October 2019.

\bibitem{keskar2016large}
N.~S. Keskar, D.~Mudigere, J.~Nocedal, M.~Smelyanskiy, and P.~T.~P. Tang.
\newblock On large-batch training for deep learning: Generalization gap and
  sharp minima.
\newblock {\em arXiv preprint arXiv:1609.04836}, 2016.

\bibitem{kim2017interpretability}
B.~Kim, M.~Wattenberg, J.~Gilmer, C.~Cai, J.~Wexler, F.~Viegas, and R.~Sayres.
\newblock Interpretability beyond feature attribution: Quantitative testing
  with concept activation vectors (tcav).
\newblock {\em arXiv preprint arXiv:1711.11279}, 2017.

\bibitem{kindermans2017learning}
P.-J. Kindermans, K.~T. Sch{\"u}tt, M.~Alber, K.-R. M{\"u}ller, D.~Erhan,
  B.~Kim, and S.~D{\"a}hne.
\newblock Learning how to explain neural networks: Patternnet and
  patternattribution.
\newblock {\em arXiv preprint arXiv:1705.05598}, 2017.

\bibitem{krizhevsky2012imagenet}
A.~Krizhevsky, I.~Sutskever, and G.~E. Hinton.
\newblock Imagenet classification with deep convolutional neural networks.
\newblock In {\em Advances in neural information processing systems}, pages
  1097--1105, 2012.

\bibitem{le2015tiny}
Y.~Le and X.~Yang.
\newblock Tiny imagenet visual recognition challenge.
\newblock {\em CS 231N}, 7, 2015.

\bibitem{Li_2020_CVPR}
T.~Li, J.~Li, Z.~Liu, and C.~Zhang.
\newblock Few sample knowledge distillation for efficient network compression.
\newblock In {\em Proceedings of the IEEE/CVF Conference on Computer Vision and
  Pattern Recognition}, June 2020.

\bibitem{lin2014microsoft}
T.-Y. Lin, M.~Maire, S.~Belongie, J.~Hays, P.~Perona, D.~Ramanan,
  P.~Doll{\'a}r, and C.~L. Zitnick.
\newblock Microsoft coco: Common objects in context.
\newblock In {\em European conference on computer vision}, pages 740--755.
  Springer, 2014.

\bibitem{liu2019structured}
Y.~Liu, K.~Chen, C.~Liu, Z.~Qin, Z.~Luo, and J.~Wang.
\newblock Structured knowledge distillation for semantic segmentation.
\newblock In {\em Proceedings of the IEEE/CVF Conference on Computer Vision and
  Pattern Recognition}, pages 2604--2613, 2019.

\bibitem{lopez2015unifying}
D.~Lopez-Paz, L.~Bottou, B.~Sch{\"o}lkopf, and V.~Vapnik.
\newblock Unifying distillation and privileged information.
\newblock {\em In {ICLR}}, 2016.

\bibitem{deepInfo}
H.~Ma, Y.~Zhang, F.~Zhou, and Q.~Zhang.
\newblock Quantifying layerwise information discarding of neural networks.
\newblock {\em In arXiv:1906.04109}, 2019.

\bibitem{mahendran2015understanding}
A.~Mahendran and A.~Vedaldi.
\newblock Understanding deep image representations by inverting them.
\newblock In {\em Proceedings of the IEEE conference on computer vision and
  pattern recognition}, pages 5188--5196, 2015.

\bibitem{menon2020distillation}
A.~K. Menon, A.~S. Rawat, S.~J. Reddi, S.~Kim, and S.~Kumar.
\newblock Why distillation helps: a statistical perspective.
\newblock {\em arXiv preprint arXiv:2005.10419}, 2020.

\bibitem{neyshabur2018a}
B.~Neyshabur, S.~Bhojanapalli, and N.~Srebro.
\newblock A {PAC}-bayesian approach to spectrally-normalized margin bounds for
  neural networks.
\newblock In {\em International Conference on Learning Representations}, 2018.

\bibitem{neyshabur2015norm}
B.~Neyshabur, R.~Tomioka, and N.~Srebro.
\newblock Norm-based capacity control in neural networks.
\newblock In {\em Conference on Learning Theory}, pages 1376--1401. PMLR, 2015.

\bibitem{papernot2016distillation}
N.~Papernot, P.~McDaniel, X.~Wu, S.~Jha, and A.~Swami.
\newblock Distillation as a defense to adversarial perturbations against deep
  neural networks.
\newblock In {\em 2016 IEEE symposium on security and privacy}, pages 582--597.
  IEEE, 2016.

\bibitem{phuong2019towards}
M.~Phuong and C.~Lampert.
\newblock Towards understanding knowledge distillation.
\newblock In {\em International Conference on Machine Learning}, pages
  5142--5151, 2019.

\bibitem{qi2017pointnet}
C.~R. Qi, H.~Su, K.~Mo, and L.~J. Guibas.
\newblock Pointnet: Deep learning on point sets for 3d classification and
  segmentation.
\newblock In {\em Proceedings of the IEEE conference on computer vision and
  pattern recognition}, pages 652--660, 2017.

\bibitem{qi2017pointnet++}
C.~R. Qi, L.~Yi, H.~Su, and L.~J. Guibas.
\newblock Pointnet++: Deep hierarchical feature learning on point sets in a
  metric space.
\newblock {\em arXiv preprint arXiv:1706.02413}, 2017.

\bibitem{rajpurkar-etal-2016-squad}
P.~Rajpurkar, J.~Zhang, K.~Lopyrev, and P.~Liang.
\newblock {SQ}u{AD}: 100,000+ questions for machine comprehension of text.
\newblock In {\em Proceedings of the 2016 Conference on Empirical Methods in
  Natural Language Processing}, 2016.

\bibitem{ribeiro2016should}
M.~T. Ribeiro, S.~Singh, and C.~Guestrin.
\newblock Why should i trust you?: Explaining the predictions of any
  classifier.
\newblock In {\em Proceedings of the 22nd ACM SIGKDD}, pages 1135--1144. ACM,
  2016.

\bibitem{ribeiro2018anchors}
M.~T. Ribeiro, S.~Singh, and C.~Guestrin.
\newblock Anchors: High-precision model-agnostic explanations.
\newblock In {\em AAAI}, volume~18, pages 1527--1535, 2018.

\bibitem{romero2014fitnets}
A.~Romero, N.~Ballas, S.~E. Kahou, A.~Chassang, C.~Gatta, and Y.~Bengio.
\newblock Fitnets: Hints for thin deep nets.
\newblock {\em arXiv preprint arXiv:1412.6550}, 2014.

\bibitem{russakovsky2015imagenet}
O.~Russakovsky, J.~Deng, H.~Su, J.~Krause, S.~Satheesh, S.~Ma, Z.~Huang,
  A.~Karpathy, A.~Khosla, M.~Bernstein, et~al.
\newblock Imagenet large scale visual recognition challenge.
\newblock {\em International journal of computer vision}, 115(3):211--252,
  2015.

\bibitem{sabour2017dynamic}
S.~Sabour, N.~Frosst, and G.~E. Hinton.
\newblock Dynamic routing between capsules.
\newblock In {\em Advances in neural information processing systems}, pages
  3856--3866, 2017.

\bibitem{selvaraju2017grad}
R.~R. Selvaraju, M.~Cogswell, A.~Das, R.~Vedantam, D.~Parikh, and D.~Batra.
\newblock Grad-cam: Visual explanations from deep networks via gradient-based
  localization.
\newblock In {\em Proceedings of the IEEE International Conference on Computer
  Vision}, pages 618--626, 2017.

\bibitem{sermanet2013overfeat}
P.~Sermanet, D.~Eigen, X.~Zhang, M.~Mathieu, R.~Fergus, and Y.~LeCun.
\newblock Overfeat: Integrated recognition, localization and detection using
  convolutional networks.
\newblock {\em arXiv preprint arXiv:1312.6229}, 2013.

\bibitem{shen2021interpretable}
W.~Shen, Z.~Wei, S.~Huang, B.~Zhang, J.~Fan, P.~Zhao, and Q.~Zhang.
\newblock Interpretable compositional convolutional neural networks.
\newblock {\em arXiv preprint arXiv:2107.04474}, 2021.

\bibitem{shwartz2017opening}
R.~Shwartz-Ziv and N.~Tishby.
\newblock Opening the black box of deep neural networks via information.
\newblock {\em arXiv preprint arXiv:1703.00810}, 2017.

\bibitem{simonyan2017deep}
K.~Simonyan, A.~Vedaldi, and A.~Zisserman.
\newblock Deep inside convolutional networks: visualising image classification
  models and saliency maps.
\newblock {\em arXiv preprint arXiv:1312.6034}, 2017.

\bibitem{simonyan2015very}
K.~Simonyan and A.~Zisserman.
\newblock Very deep convolutional networks for large-scale image recognition.
\newblock {\em In {ICLR}}, 2015.

\bibitem{simsekli2019tail}
U.~Simsekli, L.~Sagun, and M.~Gurbuzbalaban.
\newblock A tail-index analysis of stochastic gradient noise in deep neural
  networks.
\newblock In {\em International Conference on Machine Learning}, pages
  5827--5837. PMLR, 2019.

\bibitem{socher2013recursive}
R.~Socher, A.~Perelygin, J.~Wu, J.~Chuang, C.~D. Manning, A.~Y. Ng, and
  C.~Potts.
\newblock Recursive deep models for semantic compositionality over a sentiment
  treebank.
\newblock In {\em Proceedings of the 2013 conference on empirical methods in
  natural language processing}, pages 1631--1642, 2013.

\bibitem{tang2020understanding}
J.~Tang, R.~Shivanna, Z.~Zhao, D.~Lin, A.~Singh, E.~H. Chi, and S.~Jain.
\newblock Understanding and improving knowledge distillation.
\newblock {\em arXiv preprint arXiv:2002.03532}, 2020.

\bibitem{uijlings2018revisiting}
J.~Uijlings, S.~Popov, and V.~Ferrari.
\newblock Revisiting knowledge transfer for training object class detectors.
\newblock In {\em Proceedings of the IEEE Conference on Computer Vision and
  Pattern Recognition}, pages 1101--1110, 2018.

\bibitem{wah2011caltech}
C.~Wah, S.~Branson, P.~Welinder, P.~Perona, and S.~Belongie.
\newblock The caltech-ucsd birds-200-2011 dataset.
\newblock 2011.

\bibitem{wang2019dynamic}
Y.~Wang, Y.~Sun, Z.~Liu, S.~E. Sarma, M.~M. Bronstein, and J.~M. Solomon.
\newblock Dynamic graph cnn for learning on point clouds.
\newblock {\em Acm Transactions On Graphics (tog)}, 38(5):1--12, 2019.

\bibitem{weng2018evaluating}
T.-W. Weng, H.~Zhang, P.-Y. Chen, J.~Yi, D.~Su, Y.~Gao, C.-J. Hsieh, and
  L.~Daniel.
\newblock Evaluating the robustness of neural networks: An extreme value theory
  approach.
\newblock {\em arXiv preprint arXiv:1801.10578}, 2018.

\bibitem{wolchover2017new}
N.~Wolchover.
\newblock New theory cracks open the black box of deep learning.
\newblock {\em In {Quanta Magazine}}, 2017.

\bibitem{wu2019pointconv}
W.~Wu, Z.~Qi, and L.~Fuxin.
\newblock Pointconv: Deep convolutional networks on 3d point clouds.
\newblock In {\em Proceedings of the IEEE/CVF Conference on Computer Vision and
  Pattern Recognition}, pages 9621--9630, 2019.

\bibitem{wu20153d}
Z.~Wu, S.~Song, A.~Khosla, F.~Yu, L.~Zhang, X.~Tang, and J.~Xiao.
\newblock 3d shapenets: A deep representation for volumetric shapes.
\newblock In {\em Proceedings of the IEEE conference on computer vision and
  pattern recognition}, pages 1912--1920, 2015.

\bibitem{xie2017learning}
D.~Xie, T.~Shu, S.~Todorovic, and S.-C. Zhu.
\newblock Learning and inferring “dark matter” and predicting human intents
  and trajectories in videos.
\newblock {\em IEEE transactions on pattern analysis and machine intelligence},
  40(7):1639--1652, 2017.

\bibitem{xu2017information}
A.~Xu and M.~Raginsky.
\newblock Information-theoretic analysis of generalization capability of
  learning algorithms.
\newblock In {\em Advances in Neural Information Processing Systems}, pages
  2524--2533, 2017.

\bibitem{yim2017gift}
J.~Yim, D.~Joo, J.~Bae, and J.~Kim.
\newblock A gift from knowledge distillation: Fast optimization, network
  minimization and transfer learning.
\newblock In {\em Proceedings of the IEEE Conference on Computer Vision and
  Pattern Recognition}, pages 4133--4141, 2017.

\bibitem{yosinski2015understanding}
J.~Yosinski, J.~Clune, A.~Nguyen, T.~Fuchs, and H.~Lipson.
\newblock Understanding neural networks through deep visualization.
\newblock {\em arXiv preprint arXiv:1506.06579}, 2015.

\bibitem{yuan2020revisiting}
L.~Yuan, F.~E. Tay, G.~Li, T.~Wang, and J.~Feng.
\newblock Revisiting knowledge distillation via label smoothing regularization.
\newblock In {\em Proceedings of the IEEE/CVF Conference on Computer Vision and
  Pattern Recognition}, pages 3903--3911, 2020.

\bibitem{zeiler2014visualizing}
M.~D. Zeiler and R.~Fergus.
\newblock Visualizing and understanding convolutional networks.
\newblock In {\em European conference on computer vision}, pages 818--833.
  Springer, 2014.

\bibitem{zhang2016understanding}
C.~Zhang, S.~Bengio, M.~Hardt, B.~Recht, and O.~Vinyals.
\newblock Understanding deep learning requires rethinking generalization.
\newblock {\em arXiv preprint arXiv:1611.03530}, 2016.

\bibitem{zhanginterpretable}
Q.~Zhang, X.~Wang, Y.~N. Wu, H.~Zhou, and S.-C. Zhu.
\newblock Interpretable cnns for object classification.
\newblock {\em IEEE transactions on pattern analysis and machine intelligence}.

\bibitem{zhou2016learning}
B.~Zhou, A.~Khosla, A.~Lapedriza, A.~Oliva, and A.~Torralba.
\newblock Learning deep features for discriminative localization.
\newblock In {\em Proceedings of the IEEE conference on computer vision and
  pattern recognition}, pages 2921--2929, 2016.

\end{thebibliography}

\vspace{-30pt}
\begin{IEEEbiography}[{\includegraphics[width=1in,height=1.25in,clip,keepaspectratio]{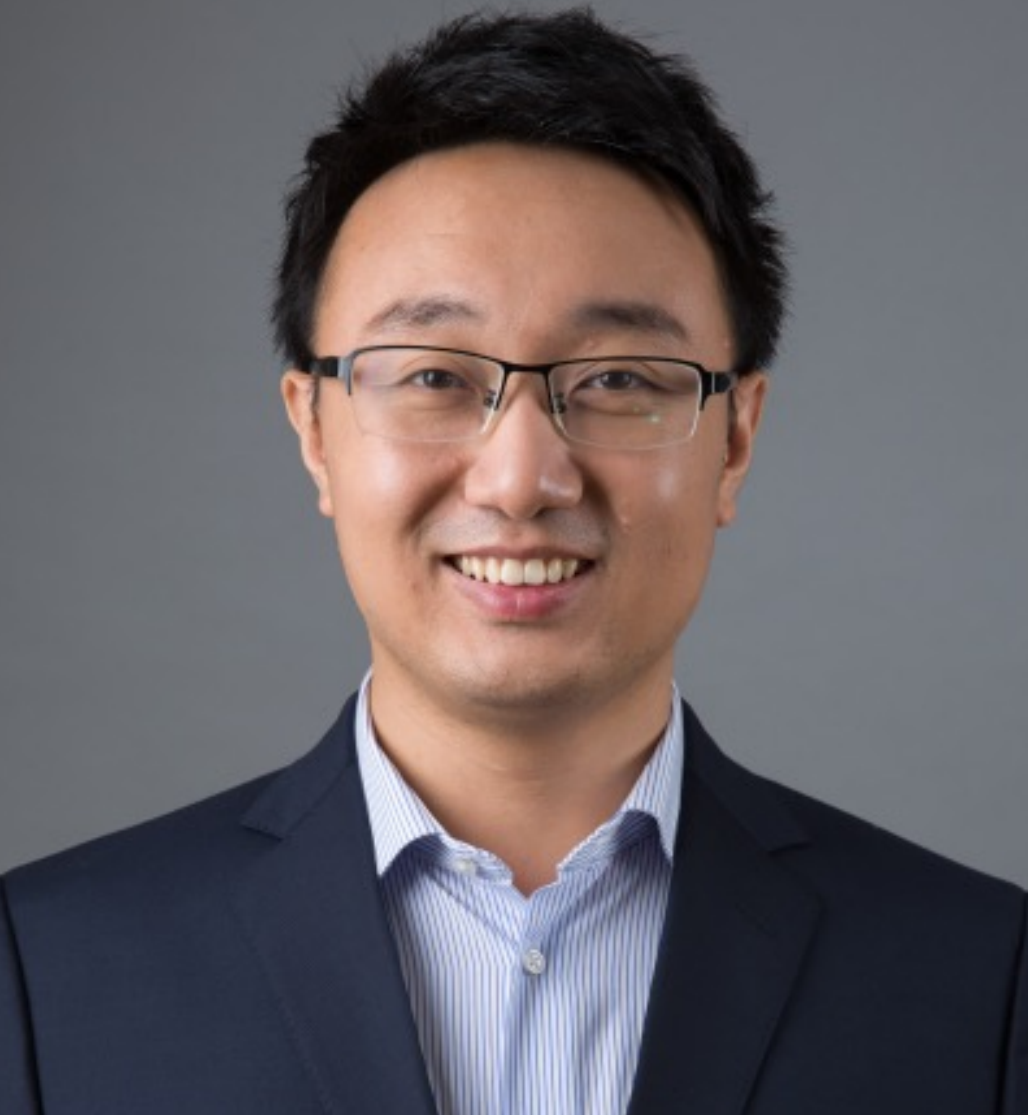}}]{Dr. Quanshi Zhang}
is an associate professor at Shanghai Jiao Tong University, China. He received the Ph.D. degree from the University of Tokyo in 2014. From 2014 to 2018, he was a post-doctoral researcher at the University of California, Los Angeles. His research interests are mainly machine learning and computer vision. In particular, he has made influential research in explainable AI (XAI). He was the co-chairs of the workshops towards XAI in ICML 2021, AAAI 2019, and CVPR 2019. He is the speaker of the tutorials on XAI at IJCAI 2020 and IJCAI 2021. He won the ACM China Rising Star Award at ACM TURC 2021. He is a member of the IEEE.
\end{IEEEbiography}
\vspace{-30pt}
\begin{IEEEbiography}[{\includegraphics[width=1in,height=1.25in,clip,keepaspectratio]{./fig/xu.pdf}}]{Xu Cheng}
is a Ph.D. student at Shanghai Jiao Tong University. Her research interests include computer vision and machine learning. She is a member of the IEEE.
\end{IEEEbiography}

\vspace{-30pt}
\begin{IEEEbiography}[{\includegraphics[width=1in,height=1.25in,clip,keepaspectratio]{./fig/lan.pdf}}]{Yilan Chen}
is a master student at University of California San Diego. His research interest is machine learning. He is a member of the IEEE.
\end{IEEEbiography}

\vspace{-30pt}
\begin{IEEEbiography}[{\includegraphics[width=1in,height=1.25in,clip,keepaspectratio]{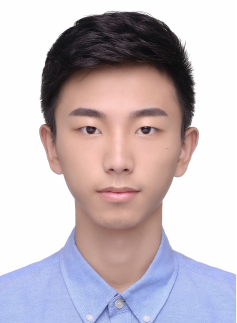}}]{Zhefan Rao}
graduated from Huazhong University of Science \&Technology. His research interest is machine learning. He is a member of the IEEE.
\end{IEEEbiography}

\vfill

\end{document}